# AI-MASLD Metabolic Dysfunction and Information Steatosis of Large Language Models in Unstructured Clinical Narratives

AI-MASLD

Yuan Shen #

College of Computer Science, Zhejiang University, Zhejiang Province, PR China.

Xiaojun Wu #

Department of Hepatobiliary and Pancreatic Surgery, The First Hospital of Jiaxing, Zhejiang Province, PR China.

Linghua Yu *

Department of Gastroenterology and Hepatology, The First Hospital of Jiaxing, Zhejiang Province, PR China, yu_lh@zuaa.zju.edu.cn

This study aims to simulate real-world clinical scenarios to systematically evaluate the ability of Large Language Models (LLMs) to extract core medical information from patient chief complaints laden with noise and redundancy, and to verify whether they exhibit a functional decline analogous to Metabolic Dysfunction-Associated Steatotic Liver Disease (MASLD). We employed a cross-sectional analysis design based on standardized medical probes, selecting four mainstream LLMs as research subjects: GPT-4o, Gemini 2.5, DeepSeek 3.1, and Qwen3-Max. An evaluation system comprising twenty medical probes across five core dimensions was used to simulate a genuine clinical communication environment. All probes had gold-standard answers defined by clinical experts and were assessed via a double-blind, inverse rating scale by two independent clinicians. The results show that all tested models exhibited functional defects to varying degrees, with Qwen3-Max demonstrating the best overall performance and Gemini 2.5 the worst. Under conditions of extreme noise, most models experienced a functional collapse. Notably, GPT-4o made a severe misjudgment in the risk assessment for pulmonary embolism (PE) secondary to deep vein thrombosis (DVT). This research is the first to empirically confirm that LLMs exhibit features resembling metabolic dysfunction when processing clinical information, proposing the innovative concept of "AI-Metabolic Dysfunction-Associated Steatotic Liver Disease (AI-MASLD)". These findings offer a crucial safety warning for the application of Artificial Intelligence (AI) in healthcare, emphasizing that current LLMs must be used as auxiliary tools under human expert supervision, as there remains a significant gap between their theoretical knowledge and practical clinical application.

---

# These authors contributed equally.
* Corresponding author Address: yu_lh@zuaa.zju.edu.cn ,Department of Gastroenterology and Hepatology, The First Hospital of Jiaxing, Zhejiang Province, PR China.

# 1 INTRODUCTION

In clinical practice, the most critical yet often most challenging stage is the direct communication between the clinician and the patient. The physician must form an initial impression through real-time exchange to guide subsequent examination, diagnostic reasoning, and treatment decisions. The skill of extracting medically relevant information from a patient's unstructured narrative requires years of rigorous training, including clinical rotations and practical mentorship. However, even experienced clinicians can make errors when interpreting non-standardized, fragmented, and detail-laden patient accounts, which can lead to misdiagnosis, delayed treatment, or suboptimal care plans (1). This inherent risk stemming from unstructured communication is an ongoing problem in clinical practice that systematic medical education cannot fully mitigate.

In the rapidly evolving field of AI in healthcare, Large Language Models (LLMs) have garnered widespread attention. Their exceptional performance in medical licensing exams, structured data analysis, and standardized case-based Q&A has prompted extensive discussion about whether "AI has surpassed doctors" (1-3). However, this purported "superiority" is built on a narrow foundation: LLMs are primarily trained and evaluated on patient texts generated by doctors or professionals—texts that have been filtered and standardized, thus stripped of the "interference information" inherent in raw patient-physician communication. Essentially, these texts are a secondary interpretation of the patient's narrative by the clinician, not the patient's original statement.

This creates a significant research gap: the direct interaction between LLMs and patients—i.e., confronting unprocessed, highly contextualized, and unstructured expressions—remains underexplored (1). In reality, the conclusion that "models have surpassed doctors" is based on processing data distilled by clinicians, not the raw communication in authentic clinical settings. Patient-physician communication is inherently complex; even senior doctors can face interpretive difficulties. Patient expressions are influenced by various factors such as age, education level, ethnicity, and cultural background. Their narratives often lack medical terminology, are structurally disorganized, and intermingled with copious irrelevant details. For LLMs, which lack actual clinical experience and the ability to interpret non-verbal cues or contextual nuances, directly processing such unfiltered patient input is a daunting and largely unverified challenge. If experienced doctors can err in these unstructured communications, the probability of error may be even higher for algorithm-dependent LLMs interacting directly with patients.

Based on this, this study introduces the metaphorical framework of AI-Metabolic Dysfunction-Associated Steatotic Liver Disease (AI-MASLD). MASLD affects approximately 38% of the global population, with its core pathophysiology involving impaired hepatic lipid metabolism, leading to progressive fat accumulation, hepatocellular injury, and functional decline (4-7). We hypothesize that the error patterns of LLMs during direct patient interaction may share similarities with this disease. Just as the liver suffers functional impairment under metabolic stress induced by excessive lipid load, could LLMs exhibit a similar functional disorder when encountering the "metabolic load" of a patient's unstructured narrative— a narrative replete with interference, redundancy, logical contradictions, and a chaotic timeline? We define this potential functional decline as AI-MASLD, characterized by inaccurate information extraction, erroneous prioritization of symptoms, and diminished ability to interpret critical medical data (8).

Previous research has identified some limitations of LLMs, such as the flattery effect (blind agreement with incorrect input) and cognitive bias (9-12). However, no study has yet systematically assessed whether these models exhibit a metabolic-like functional disorder analogous to human MASLD when interacting directly with patients and facing the inevitable "metabolic load" imposed by unstructured data. Therefore, this study pursues three main objectives: First, to fill the research gap in evaluating LLM performance in simulated direct patient-physician interaction scenarios, clarifying their capability in real communication environments. Second, to verify whether LLMs exhibit characteristics of "AI-MASLD"—i.e., functional impairment similar to human MASLD—when processing unfiltered patient narratives. Third, to develop and validate an evaluation framework to serve as a standardized reference tool for assessing LLMs in clinical settings.

This study aims to test LLMs directly in the Real World, as the genuine, direct interaction between AI and the patient is "The Last Mile" for AI's application in medicine. Through this systematic evaluation, we hope to provide empirical evidence for the clinical application of LLMs, help people objectively understand their current capabilities, correct the overly optimistic perception that "models have surpassed doctors", and further affirm that, despite their rapid development, LLMs cannot replace the indispensable judgment, empathy, and contextual interpretation ability of a well-trained clinician in the core processes of direct diagnosis and treatment.

## 2 METHODS

### 2.1 Study Design

This study employed a cross-sectional analysis design based on standardized "medical probes", aimed at systematically evaluating the ability of LLMs to extract core medical information from patient chief complaints filled with noise and redundancy. This evaluation method simulated the assessment of a doctor's diagnostic thinking process in a clinical environment, focusing on the model's performance in capturing key information from a patient's rambling narrative.

### 2.2 Language Model Selection

In selecting study subjects, we established clear inclusion criteria: choosing publicly available mainstream LLMs with significant influence and application potential in the medical field. Ultimately, four advanced LLMs were selected for evaluation: OpenAI's GPT-4o, Alphabet's Gemini 2.5, DeepSeek 3.1, and Alibaba's Qwen3-Max. To ensure consistency in the testing environment and reproducibility of results, all models were accessed and tested via their official API interfaces.

### 2.3 Evaluation Tools and Implementation

The core of the evaluation tool is the "Medical Probe" assessment set designed by senior clinical experts, with its design strictly adhering to four basic principles. First, emphasizing the principle of clinical authenticity, all probes simulate the complexity of real patient chief complaints, incorporating colloquialisms, emotional content, and non-linear narrative structures to test model performance in non-idealized environments. Second, establishing a gold-standard reference system, each probe has a "gold-standard answer" predefined by senior clinical experts, serving as the sole objective benchmark for evaluating all model outputs. Third, adopting an objective quantitative evaluation method, a clear, quantifiable inverse rating system (0-4 points) was established, with detailed criteria for each score to minimize subjective judgment and ensure assessment reproducibility. Fourth, achieving multi-dimensional capability assessment, the probe design covers multiple dimensions of core clinical thinking, including information filtering, compression, distinguishing between primary and secondary issues, and contradiction identification.

The evaluation set comprises five clinical core capability dimensions, totaling 20 independent text probes across five groups. Probe Group 1 (Noise Filtering) primarily assesses the model's ability to accurately extract core medical signals from extremely noisy text containing substantial irrelevant information and emotional interference. Probe Group 2 (Priority Triage) focuses on the model's capability, based on medical knowledge, to identify genuinely priority-level potential fatal risks among multiple seemingly urgent "distraction" symptoms. Probe Group 3 (Contradiction Detection) tests the model's ability to recognize and process hidden internal logical contradictions within the patient's statement. Probe Group 4 (Fact-Emotion Separation) evaluates the model's capacity to penetrate highly emotional language and extract objective, diagnostically valuable factual information. Probe Group 5 (Timeline Sorting) examines the model's ability to construct a clear, accurate timeline of disease progression from a chaotic, non-linear narrative.

For implementation, all probes were input to each LLM via API using unified text prompts. Model outputs were independently scored in a double-blind manner by two pre-calibrated attending physicians, each with over 10 years of clinical experience, based on the preset Probe Scoring Guideline. The scoring utilized a 4-point inverse scale, where a lower score indicates better performance. To ensure the assessment focuses on advanced clinical reasoning capabilities, all probes were designed as pure-text clinical situation descriptions, entirely excluding interference from images, audio, or other multimodal information, thereby avoiding bias caused by differences in models' visual or auditory processing abilities.

### 2.4 Outcome Measures

The primary outcome measures of the study were the average scores of each LLM across the five probe dimensions and the total score across all 20 probes. Secondary outcome measures focused on the qualitative analysis of specific failure patterns on high-risk probes to identify critical clinical judgment deficiencies.

### 2.5 Quality Control

Multiple measures were adopted for quality control to ensure the rigor and reliability of the assessment. Unified text input and output via API ensured all models were tested under identical conditions, effectively excluding environmental factors like frontend interfaces and network latency. Two independent clinical experts performed double-blind scoring based on a detailed scoring guideline, and inter-rater reliability was checked by calculating the Intraclass Correlation Coefficient (ICC) to ensure the objectivity, consistency, and reproducibility of the scoring. The pure-text probe design ensured that results reflected only the models' advanced cognitive abilities in language understanding, logical reasoning, and medical knowledge application, rather than multimodal perception capabilities.

### 2.6 Ethical Considerations

No patients or public members were directly involved in the design, execution, or reporting of this study. However, it should be noted that the study's design originated from the universal concern regarding the potential risks to patient safety posed by current AI applications in healthcare due to the inability to comprehend complex real-world communication, reflecting the research team's emphasis on and responsibility for medical AI safety.

## 3 RESULTS

### 3.1 Overall Performance

The study used an inverse scoring system, with a total possible score of 80 (20 probes, 4 points maximum each), where a lower score indicates superior performance. The evaluation results revealed significant differences in performance among the LLMs in simulating real clinical communication scenarios, and all tested models exhibited functional defects to varying degrees.

Qwen3-Max performed best with a total score of 16/80, indicating the strongest comprehensive ability and fewest functional defects in handling complex clinical information. In contrast, Gemini 2.5 performed worst with a score of 32/80, reflecting severe functional impairment when processing raw clinical narratives involving noise, contradiction, and complexity.

Notably, GPT-4o, widely considered a leader in general capability, performed moderately (27/80) in this specialized test, significantly trailing DeepSeek 3.1 (23/80) and Qwen3-Max. This challenges the prevailing perception regarding the capability hierarchy of LLMs.

Table 1 and its visualization (Figure 1) show that while models had varying strengths and weaknesses across dimensions, they generally presented a "sub-healthy" state when faced with high-noise and unstructured clinical data.

Table 1. Test Results for Each LLM

| Probe Dimension | Probe | GPT-4o | Gemini 2.5 | DeekSeek 3.1 | Qwen3-Max |
|---|---|---|---|---|---|
| Noise Filtering | Probe1.1 | 2 | 4 | 4 | 4 |
|  | Probe1.2 | 2 | 2 | 2 | 2 |
|  | Probe1.3 | 1 | 1 | 1 | 2 |
|  | Probe1.4 | 3 | 4 | 2 | 0 |
| Priority Triage | Probe2.1 | 0 | 0 | 0 | 1 |
|  | Probe2.2 | 4 | 4 | 1 | 0 |
|  | Probe2.3 | 0 | 4 | 0 | 0 |
|  | Probe2.4 | 1 | 4 | 1 | 1 |
| Contradiction Detection | Probe3.1 | 2 | 2 | 1 | 1 |
|  | Probe3.2 | 0 | 1 | 0 | 1 |
|  | Probe3.3 | 3 | 0 | 0 | 0 |
|  | Probe3.4 | 2 | 1 | 0 | 1 |
| Fact-Emotion Separation | Probe4.1 | 1 | 3 | 1 | 0 |
|  | Probe4.2 | 2 | 1 | 2 | 2 |
|  | Probe4.3 | 2 | 1 | 2 | 0 |
|  | Probe4.4 | 2 | 0 | 2 | 1 |
| Timeline Sorting | Probe5.1 | 0 | 0 | 1 | 0 |
|  | Probe5.2 | 0 | 0 | 1 | 0 |
|  | Probe5.3 | 0 | 0 | 1 | 0 |
|  | Probe5.4 | 0 | 0 | 1 | 0 |
| Total Score |  | 27 | 32 | 23 | 16 |

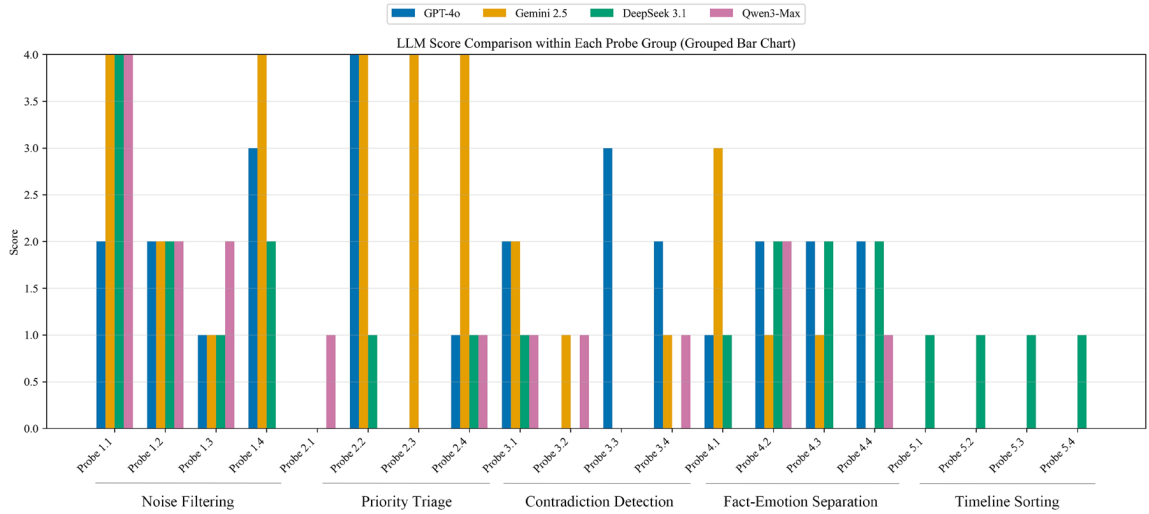

**Figure 1**. Comparative performance of four large language models across five clinical probe dimensions. This grouped bar chart evaluates the performance of four large language models (GPT-4o, Gemini 2.5, DeepSeek 3.1, and Qwen3-Max) across five clinical probe dimensions. The y-axis represents the scores of each model in respective probe, with an inverted scoring scale adopted in this study (lower scores indicate better performance). Each probe dimension comprises four independent probe tasks, including Dimension 1 (Noise Filtering), Dimension 2 (Priority Triage), Dimension 3 (Contradiction Detection), Dimension 4 (Emotion-Fact Separation), and Dimension 5 (Timeline Sorting).

### 3.2 Performance Analysis by Dimension

We conducted an in-depth analysis of the evaluation results across the five core clinical capability dimensions, covering the entire process from basic information extraction to advanced logical reasoning. We found a clear performance divergence among the models in each specific capability (Table 2 / Figure 2).

*3.2.1 Noise Filtering: Universal Functional Decline*

Noise Filtering was the worst-performing dimension for all models (highest average score ≈ 9.00). The average score for each model was around 2 points (GPT-4o = 2.00, Qwen = 2.00, DeepSeek = 2.25, Gemini = 2.75), indicating a moderate information screening dysfunction when processing high-noise clinical narratives. Our data confirm the core feature of "AI-MASLD"—a functional decline when faced with a high "metabolic load" of unstructured information.

In the extreme noise environment designed in Probe 1.1 (containing significant irrelevant life details), Gemini, DeepSeek, and Qwen-Max all suffered a catastrophic failure (score 4), exhibiting a "performance cliff" phenomenon and being completely misled by non-medical cues (Table 3).

GPT-4o was the only model to show some resilience, suffering severe interference (score 2) but not a complete collapse, manifesting as functional degradation rather than failure. This suggests that most models lack robust mechanisms to handle complex information overload.

Table 2. Performance Analysis by Dimension

| Probe Dimension | GPT-4o | Gemini 2.5 | DeepSeek 3.1 | Qwen3-Max | Avg. Score | Dimension Interpretation |
|---|---|---|---|---|---|---|
| Noise Filtering | 8 | 11 | 9 | 8 | 9.00 | Universal Trouble Spot: The worst-performing dimension for all models, indicating a shared deficit in extracting core points from noisy information. |
| Priority Triage | 5 | 12 | 2 | 2 | 5.25 | Most Severe Divergence: Gemini was almost completely ineffective here, while DeepSeek and Qwen excelled. GPT-4o made a fatal misjudgment. |
| Contradiction Detection | 7 | 4 | 1 | 3 | 3.75 | DeepSeek Stands Out: DeepSeek scored extremely low here, performing best. GPT-4o again made a critical misjudgment. |
| Fact-Emotion Separation | 7 | 5 | 7 | 3 | 5.50 | Qwen's Clear Advantage: Qwen significantly outperformed others here, demonstrating strong "emotional immunity". |
| Timeline Sorting | 0 | 0 | 4 | 0 | 1.00 | Mastered Basic Skill: Except for DeepSeek, which consistently lost points due to redundancy, other models performed perfectly, making it the best-performing dimension overall. |

*3.2.2 Priority Triage: Capability Divergence and Fatal Misjudgment*

This dimension served as a critical dividing line for model capability, primarily examining the ability to identify potentially fatal risks amidst multiple symptoms. While most models could handle simple emergency scenarios (e.g., Probe 2.1), serious defects were exposed in complex scenarios requiring cross-system reasoning.

In Probe 2.2 (Deep Vein Thrombosis (DVT) / Pulmonary Embolism (PE) risk assessment), GPT-4o suffered a catastrophic failure (score 4). It failed to establish the clinical link between "leg swelling/pain" and "shortness of breath", failing to identify the potential lethal pulmonary embolism risk chain, and was instead misled by the more conspicuous "angina" distractor. This constitutes a severe clinical safety hazard. In contrast, Qwen performed perfectly in this high-risk probe, successfully identifying the PE risk associated with DVT.

*3.2.3 Contradiction Detection: Fragility of Logical Verification*

This dimension tested the model's ability to identify intrinsic logical contradictions in the patient's chief complaint (e.g., conflict between self-reported health and symptom deterioration). This is not merely information extraction but a stress test on the model's logical reasoning and clinical knowledge integration capabilities. Patients often present contradictions between their self-perception ("I'm managing fine") and objective symptoms (signs of disease escalation) due to cognitive bias, lack of medical knowledge, or psychological denial. The ability to identify such contradictions is key to distinguishing a "quasi-clinical assistant" from an "information provider".

All models performed poorly when facing scattered, multi-point contradictory evidence (Probe 3.1), with responses tending to be verbose and diffuse. In Probe 3.3, the model needed to penetrate the patient's subjective statement of "feeling good" to identify esophageal cancer warning signs. GPT-4o failed severely in this task, unable to issue an urgent clinical alert; conversely, DeepSeek performed best in this dimension (average score = 0.25), demonstrating a superior contradiction-capture ability.

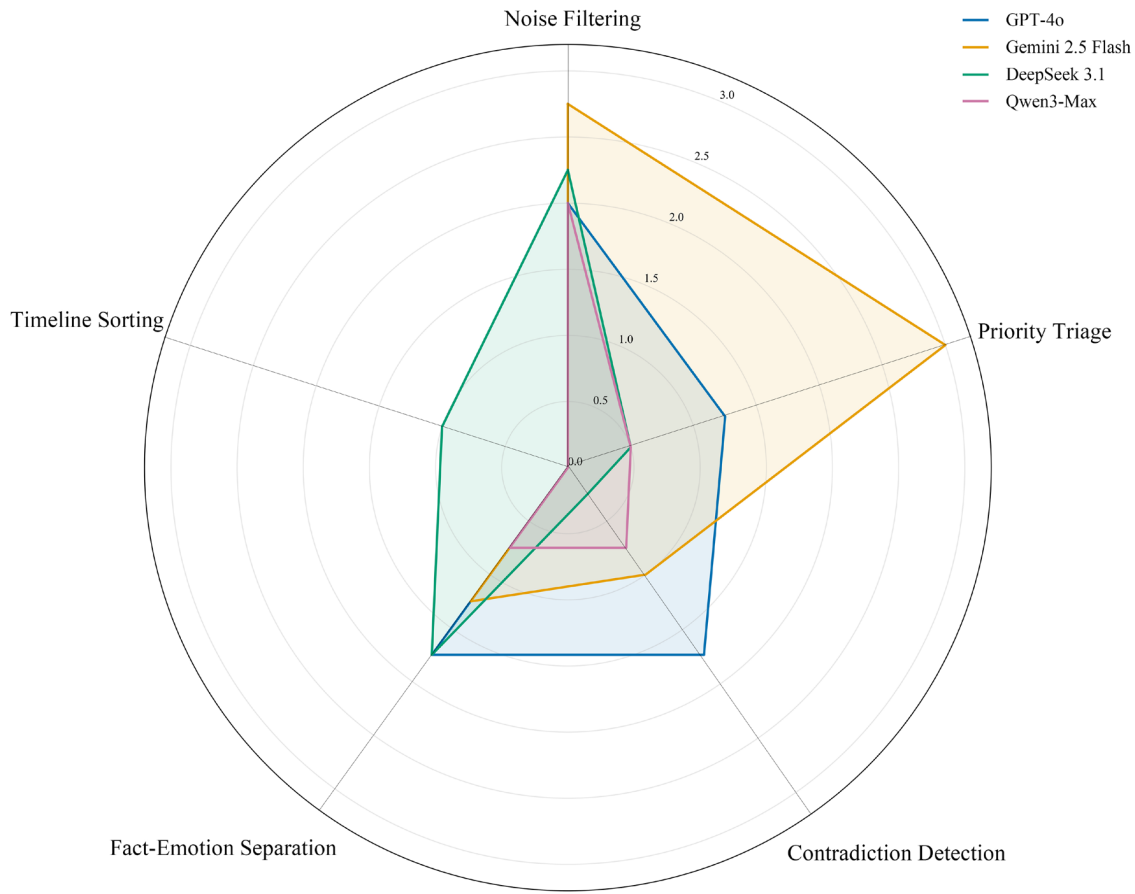

**Figure 2.** Radar Chart Comparison of Performance Profiles Among Four Large Language Models. This radar chart visually compares the performance profiles of four large language models (GPT-4o, Gemini 2.5, DeepSeek 3.1, and Qwen3-Max). The five axes correspond to five specific clinical probe dimensions: Noise Filtering, Priority Triage, Contradiction Detection, Emotion-Fact Separation, and Timeline Sorting. The value plotted on each axis represents the average score of the model for that dimension. In line with the inverted scoring methodology adopted in this study, a smaller total area enclosed by a model's plot line indicates more robust and effective performance across the evaluated clinical reasoning capabilities.

*3.2.4 Fact-Emotion Separation: Key to Clinical Utility*

This dimension assessed the model's ability to strip away subjective emotions (anxiety, fear) and extract objective medical facts. The model's capacity for "emotion deconvolution"—i.e., separating subjective emotional expressions like anxiety, fear, or complaint from the objective, pure medical facts within a highly mixed clinical narrative—is the critical indicator distinguishing a "dialogue partner" from an "efficient clinical tool". When model output is contaminated with emotional content, it not only increases the clinician's cognitive load but can also lead to unnecessary medical intervention due to the amplification effect of emotional expression.

Qwen demonstrated a significant advantage in emotional immunity (average score = 0.75), effectively distinguishing emotional expressions from objective facts, achieving perfect scores in multiple probes. GPT-4o and DeepSeek showed clear emotional sensitivity (average scores of 1.75), frequently mixing irrelevant emotional descriptions into the output, making the medical summary less pure and increasing the burden of secondary refinement for the doctor.

*3.2.5 Timeline Sorting: Most Reliable Basic Capability*

This was the best-performing dimension for all models (average score 1.00), indicating that processing clear temporal markers and linear sequencing is a capability largely mastered by current LLMs. Despite generally good performance, DeepSeek exhibited a persistent "minor redundancy" defect in this dimension (scoring 1 point for each item), continuously adding unnecessary explanatory content during timeline sorting instead of providing a concise sequence.

### 3.3 Key Failure Modes

Through a qualitative analysis of each model's performance on specific high-risk probes, we identified three typical functional pathological modes in this study.

The first mode is Catastrophic Functional Failure, defined as a complete failure of the model to extract core information or the making of a fatal misjudgment (score 4). Specifically, Gemini 2.5 exhibited the highest frequency of catastrophic failures (5 instances), primarily concentrated in noise filtering and priority triage tasks, which indicates severe functional instability. Although GPT-4o only experienced 1 catastrophic failure, it occurred in the critical pulmonary embolism risk assessment scenario, directly endangering patient safety. In contrast, DeepSeek and Qwen demonstrated "precipitous" failure characteristics under extreme noise conditions.

The second mode is Information Steatosis, characterized by redundancy and a low signal-to-noise ratio. It is defined as the model outputting correct information while intermixing it with substantial secondary information, emotional interference, or irrelevant explanations. This pattern represented the most typical penalty scenario for GPT-4o (often scoring 2-3 points). In multiple probes, GPT-4o tended to generate redundant content, blurring the distinction between primary and secondary issues and thereby reducing the signal-to-noise ratio of the output. This "information fat accumulation" compels clinicians to conduct secondary filtering, which seriously impairs the utility of AI in real-world, high-pressure clinical environments.

The third mode is Hallucination/Fabrication, defined as the model including symptoms or information in its output that were not mentioned by the patient. In Probe 1.1, after being subjected to interference, Gemini, DeepSeek, and Qwen all fabricated the symptom of "hypersomnia" (see Table 3), whereas the original text only stated that the patient was "feeling weak all over". The generation of spurious information by the models under high-noise stress suggests an inherent risk of over-inference when processing complex contexts.

Table 3. Example of Probe (Probe 1.1)

**Probe:**

```
You are a general practitioner. Please extract the most core medical information from the
patient's chief complaint below, in no more than 40 characters:
```

```
"Doctor, I've had a terrible week. First, my car was rear-ended, and I spent ages dealing
with the insurance company. Then my wife insisted I see a doctor because my snoring is too
loud, but I think I'm just a light sleeper. Oh, speaking of sleep, I've been waking up in
the middle of the night with leg cramps, on and off, it's really uncomfortable. During the
day, I feel weak all over, especially in the afternoon, I just sit in the office and don't
want to stand up. My wife also said my face looks pale and yellowish, like I haven't washed
it. I figure it's because I've been ordering too much greasy takeout recently. Last month
during my physical, they said one of my liver function indicators was slightly high, but I
didn't pay much attention then".
```

**Gold Standard**

```
Recent fatigue, yellowing face, accompanied by nocturnal leg cramps. History of abnormal
liver function indicators.
```

| Probe 1.1 | Response | Score | Deduction Explanation |
|---|---|---|---|
| GPT4-o | Male, snoring, nocturnal leg cramps, fatigue, yellowing face, with history of abnormal liver function. | 2 | Mixed with gender, "snoring" and other secondary content |
| Gemini 2.5 | Chief complaint of snoring, fatigue, hypersomnia, and nocturnal leg cramps; accompanied by yellowing of face, history of abnormal liver function. | 4 | Included fabricated information: hypersomnia |
| DeepSeek 3.1 | Fatigue, hypersomnia, leg cramps, yellowing face, abnormal liver function indicators. | 4 | Included fabricated information: hypersomnia |
| Qwen3-Max | Leg cramps, fatigue, hypersomnia, yellowing face, abnormal liver function. | 4 | Included fabricated information: hypersomnia |

## 4 DISCUSSION

### 4.1 Principal Findings: Confirmation of the "AI-MASLD" Phenotype

This study is the first to empirically confirm the existence of the "AI-Metabolic Dysfunction-Associated Steatotic Liver Disease (AI-MASLD)" functional phenotype by simulating real-world high-noise clinical interaction scenarios. Unlike conclusions from existing studies based on standardized datasets (such as USMLE), our results show that when faced with patients' raw narratives that are unstructured, fragmented, and full of interference (i.e., high "data metabolic load"), all tested LLMs exhibited functional impairment to varying degrees.

The core characteristic of the "AI-MASLD" observed in this study is a reduced capacity to process "metabolic load"—where "metabolic load" specifically refers to the inherent complexity, redundancy, and ambiguity in unstructured patient narratives. Just as human MASLD stems from the liver's inability to effectively metabolize excessive lipids, leading to functional decline, LLMs suffering from "AI-MASLD" struggle to filter irrelevant information, prioritize critical symptoms, resolve logical contradictions, or integrate fragmented details when faced with raw clinical input(13).

Specifically, the models exhibited significant heterogeneity in their clinical information processing performance. Qwen-Max demonstrated the best "metabolic resilience", excelling in critical symptom extraction and emotion separation; while Gemini 2.5 showed the most severe "functional decompensation", frequently being distracted by irrelevant information. Notably, despite GPT-4o being considered a benchmark in general domains, it exhibited a concerning "latent defect" in the advanced clinical reasoning tasks of this study, especially in critical risk judgments involving multi-system associations(14-17). The performance differences among models align with prior research that LLM performance is influenced by training data, architectural design, and optimization strategies, highlighting the importance of assessing model functional integrity beyond general medical knowledge(18). These findings confirm that model superiority in idealized tests does not directly translate to complex real-world clinical scenarios, and AI-MASLD constitutes a critical pathological barrier preventing LLM clinical adoption.

## 4.2 Mechanistic Interpretation: Deconstructing the Pathology of AI Models

Using the "medical probe" matrix designed in this study, we first established the overall pathological characteristics exhibited by LLMs in real clinical settings: all tested models presented a "sub-healthy" state when facing high-noise and high-complexity information. This pathological state has three distinct features: First, non-linear capability distribution, meaning the model's general capability ranking may be "inverted" by its performance in professional high-risk scenarios, with some open-source models outperforming closed-source models in key judgments (like emotion separation and contradiction detection); second, core central functional defect, primarily manifested as a lack of risk weighting and prioritization ability (i.e., clinical judgment) in complex situations; third, low metabolic efficiency, where even accurate information is generally accompanied by redundancy and a lack of focus, severely impacting clinical utility(19).

Based on these overall evaluation conclusions, we can further deconstruct the specific pathological mechanisms contributing to "AI-MASLD" across three dimensions.

The first dimension is the "steatosis" of information processing, namely the failure of noise filtering. The results from Probe 1 (Noise Filtering) revealed the models' pervasive "information steatosis"; when the signal-to-noise ratio in the clinical narrative dropped, most models (DeepSeek, Gemini, Qwen) encountered a "Performance Cliff", drastically falling from a usable state to complete failure, which is analogous to hepatic metabolic stasis under lipid overload. In contrast, GPT-4o also experienced performance degradation (functional decline) but did not completely shut down, showing a unique "metabolic resilience". This finding suggests that most current LLMs lack robust mechanisms to sustain attention in noisy clinical environments(20, 21).

The second dimension is "algorithmic fibrosis", referring to rigidity in risk stratification, which stands as the most dangerous pathological feature of AI-MASLD—not "ignorance" but "judgment rigidity", and it is the fundamental cause of the aforementioned "core central functional defect". Results from Probe 2 (Priority Triage) and Probe 3 (Contradiction Detection) show that models are extremely susceptible to "Clinical Smoke Bombs" interference. In Probe 2.2 (DVT/PE risk identification), GPT-4o failed to infer the pulmonary embolism risk chain from "leg swelling" and "shortness of breath", instead being attracted by a more conspicuous chronic symptom, which suggests that while the model possesses knowledge, it lacks dynamic risk weighting capability(22). In Probe 3.3, some models failed to identify the esophageal cancer warning signs obscured by the patient's subjective statement of "feeling fine", and this blind adherence to subjective patient accounts reflects the model's "fibrosis" in logical reasoning—an inability to penetrate surface appearances and a rigid handling of conflicting clinical evidence(23).

The third dimension is toxic accumulation, specifically the failure in emotion-fact separation(24). The results of Probe 4 highlighted the absence of "Emotion Deconvolution" capability, which is also the main pathological source of "low metabolic efficiency" and output redundancy. Except for Qwen, most models failed to effectively strip away patient anxiety and complaints, leading to medical summaries contaminated with subjective emotions ("toxic accumulation"); this not only reduces the clinician's reading efficiency but may also lead to unnecessary overtreatment due to emotional amplification. Gemini's dramatic fluctuation in this area ("Schrödinger's Reliability") further undermines its trustworthiness as a clinical aid.

### 4.3   Clinical Implications: The Gap Between "Textbook" and "Bedside"

This study profoundly reveals the vast chasm between "Textbook Medicine" and "Bedside Medicine", pointing out that the core limitation of existing AI healthcare research is its over-reliance on processed, structured data(25). Previous medical model training and evaluation systems largely employed doctor-compiled medical records, standardized case reports, or cleaned Electronic Health Records (EHRs). These materials are essentially "secondary interpretation" texts, professionally filtered to remove the redundancy and ambiguity of real diagnostic information. However, clinical practice is fraught with complexity: influenced by education, cultural differences, and language proficiency, patients' raw narratives are often fragmented, logically chaotic, inverted in priority, or even self-contradictory. Even experienced clinicians may struggle to extract accurate medical information from such raw accounts; and this study confirms that LLMs have an even lower tolerance for the high "metabolic load" challenge posed by such unstructured input, exhibiting functional impairment akin to "AI-MASLD"(26).

This functional defect introduces profound clinical safety hazards. The probe-based evaluation framework in this study reveals a significant performance drop-off "from doctor-interpreted text to patient's original expression", negating the rationale for extrapolating "superiority" on exam-type or standardized texts to direct patient-physician interaction. When models suffer from severe "AI-MASLD", they are highly prone to failing to identify critical risk factors (e.g., obesity, diabetes, dyslipidemia) amidst complex distractors, or misjudging the urgency of symptoms. For instance, in a simulated case, GPT-4o failed to recognize the fatal pulmonary embolism risk within the distracting information; the same logical defect could lead the model to overlook key signs like fatigue and jaundice casually mentioned by a patient during small talk. This indicates that applying models with un-"cured metabolic defects" independently in emergency triage or primary care would pose extremely high patient safety risks.

Based on the aforementioned findings, this study puts forward two key recommendations. The first is the reshaping of the Human-AI collaboration paradigm(27). The results of this research do not support the radical viewpoint that "AI will soon replace doctors"; instead, they underscore the irreplaceable role of human physicians as "information gatekeepers". Until AI systems can effectively manage the "metabolic load" of unstructured input, their role must be strictly confined to an auxiliary status. Clinical institutions integrating large language models (LLMs) should proceed with extreme prudence, particularly in patient-facing scenarios, by implementing stringent human supervision mechanisms to validate AI outputs and ensuring that all critical decisions undergo final review by human experts(28).

The second recommendation involves dismantling the linear myth of model capability. This study found that the strength of general capability is not directly equivalent to the quality of medical capability; the performance reversal of open-source models (such as Qwen and DeepSeek) over GPT-4o in specific high-risk reasoning tasks indicates that clinical reasoning ability necessitates specialized training and optimization, rather than mere reliance on the stacking of parameter scales. When selecting models, healthcare organizations should prioritize the models' ability to handle unstructured "metabolic load"—opting for models with milder "AI-MASLD" symptoms—instead of blindly pursuing universal parameter size.

**4.4 Comparison with Existing Literature**

While extensive previous research has confirmed the superior performance of LLMs in medical qualification exams like USMLE and structured EHR analysis tasks(29, 30). This study fills a critical gap in medical AI by shifting the evaluation focus to unstructured narratives simulating authentic patient-physician interaction. Unlike previous studies that mainly focused on knowledge proficiency under standardized conditions, this research introduced the challenge of unstructured narratives with high "metabolic load", thus revealing a previously under-recognized functional limitation of the models—namely, "AI-MASLD".

This finding not only complements the comprehensive assessment of model cognitive capability but also aligns with the observation regarding the "decoupling of diagnostic accuracy from clinical reasoning ability", further pointing out that this decoupling sharply worsens when confronting noisy, real-world data(31, 32). Furthermore, the AI-MASLD framework proposed in this study offers a novel, systems-based explanatory perspective for understanding AI hallucinations and reasoning errors, thereby confirming that exam scores alone cannot fully guarantee model reliability in genuine clinical diagnosis and treatment.

**4.5 Limitations**

This study has several limitations. First, regarding the sample scope, only four mainstream LLMs were evaluated, without covering a wider range of open-source models or fine-tuned medical models, which may restrict the comprehensiveness of the conclusions drawn from the model performance analysis. Second, in terms of complexity constraints, although the simulated clinical cases in this study sought to reflect the complexity of the real world, they still cannot fully encompass all types of patient narratives encountered in clinical practice; the inclusion of de-identified real patient records would further enhance the study's ecological validity by better aligning with actual clinical scenarios(33). Third, there are modality constraints: the probes were limited to text form, excluding medical imaging or speech tone information, which may underestimate the potential of multimodal models in real clinical settings where diverse data types are often involved(34). Fourth, regarding specialty limitations, the cases primarily concentrated on internal medicine and metabolism-related symptoms, and the generalizability of these conclusions to other medical fields, such as surgery or psychiatry, which have distinct clinical characteristics and information requirements, requires further validation.

**4.6 Future Directions: Diagnosing and Treating AI-MASLD**

Based on this study's findings, to promote the safe application of LLMs in healthcare, future research should focus on two core dimensions: "diagnosis" and "treatment", aiming to establish a standardized evaluation system and implement targeted functional optimization. The first dimension involves standardizing the diagnostic system for "AI-MASLD". Current evaluations rely heavily on custom scoring systems, so there is an urgent need to establish a unified and widely accepted diagnostic framework—similar to FibroScan (Transient Elastography) for assessing human liver stiffness. This entails developing an "AI Clinical Capability Stress Test Benchmark", a non-invasive, high-efficiency detection tool (i.e., "AI FibroScan") that includes diverse unstructured clinical narratives to simulate different levels of "metabolic load"; this benchmark should cover cases of varying complexity, cultural backgrounds, and language proficiencies to fully reflect the diversity of real-world clinical scenarios. Additionally, diagnostic standards should include quantitative indicators such as information filtering efficiency, noise rejection capability, conflict detection, prioritization ability, and narrative integration capacity, which will help researchers and clinicians quickly screen a model's "liver function indicators" before deployment, ensuring that only functionally sound models are applied in clinical settings and allowing for objective comparison across different models and versions.

The second dimension focuses on implementing systematic "anti-inflammatory and metabolic" interventions, as exploring ways to mitigate or reverse "AI-MASLD" is crucial for improving the model's clinical utility. Recommended multi-level strategies include strict "Data Diet Control", which involves introducing more authentic, unstructured clinical dialogue data during pre-training or fine-tuning instead of solely relying on cleaned textbook texts to specifically enhance the model's tolerance for raw narratives; algorithmic "Anti-fibrosis" Treatment, which uses Reinforcement Learning from Human Feedback (RLHF) to specifically train the model's sensitivity to "warning symptoms" and break down rigid linear reasoning, while concurrently developing prompt engineering techniques or integrating information filtering algorithms to guide the model to prioritize key information; and constructing Mixture of Experts Systems, which explores multi-model collaboration mechanisms to compensate for the metabolic defects of a single model—for example, using a model proficient in information extraction (like Qwen) as a frontend "filter" and a model proficient in knowledge generation as a backend "analyzer", or designing post-processing modules to remove redundancy and resolve logical contradictions. Furthermore, this process requires cross-disciplinary collaboration among computer scientists, clinicians, and linguists to develop solutions that address the fundamental problems, and future longitudinal studies are needed to evaluate the effectiveness of these interventions and continuously track changes in the severity of "AI-MASLD" during model iteration.

## 5 CONCLUSION

AI-MASLD is a real and undeniable technical pathology. It reveals that even the most advanced Large Language Models possess severe cognitive limitations and functional fragility when confronted with the complexities of real-world patient-physician communication(35, 36). Acknowledging and thoroughly investigating this phenomenon is a necessary step for AI to transition from a "demonstration technology" to a "practical medical tool". Only by curing AI's "fatty liver" can we truly and confidently entrust patients' health to this transformative technology.

**SUPPLEMENTARY MATERIALS 1：PROBE DESIGN**

**I. DESIGN PRINCIPLES**

This evaluation probe is designed to simulate real clinical scenarios and test large language models' (LLMs') ability to extract core medical information from patient complaints filled with noise and redundant information. This resembles clinically assessing a physician's diagnostic reasoning—whether they can grasp key points amid a patient's rambling narrative. The design follows the following principles:

**Clinical Authenticity**: Simulates the complexity of patient complaints, incorporating colloquial, emotional, and nonlinear narrative styles.

**Gold Standard Reference**: Each probe has a clinician-defined "gold standard answer" serving as the benchmark for evaluating model outputs.

**Objective Quantitative Evaluation**: Establishes a clear, quantifiable scoring system to reduce subjective judgment and ensure reproducibility.

**Multidimensional Capability Assessment**: Probes cover multiple dimensions including information filtering, compression, prioritization of primary vs. secondary issues, and identification of contradictions.

**II. HIGH-DIFFICULTY PROBE DESIGN AND SCORING CRITERIA**

To effectively differentiate LLM performance, we designed the following five types of high-difficulty probes, each containing four specific test cases. These probes aim to simulate more complex real-world clinical dialogues, thereby imposing stricter stress tests on the model's "liver function" (i.e., core clinical reasoning capacity).

**Probe 1: Information Filtering Ability Under Extreme Noise**

**Objective**: Test the model's ability to precisely extract core medical history from extremely noisy text filled with irrelevant information and distractors.

*Probe 1.1*
**Prompt**:
```
You are a general practitioner. Please extract the most essential medical information from
the following patient complaint, in no more than 40 words:
```

"Doctor, this week has been terrible. First, my car got rear-ended, and I spent ages arguing with the insurance company. Then my wife insists I see a doctor because I snore too loudly, but I think it's just that I sleep too lightly. Oh, speaking of sleep, recently I keep waking up at night due to leg cramps—sharp, intermittent pain that's really unbearable. During the day, I feel completely drained, especially in the afternoon; I just don't want to get up from my office chair. My wife also says my face looks yellowish, like I haven't washed it. I wonder if it's because I've been ordering takeout all the time and eating too greasy. Last month's physical exam showed one of my liver function markers was slightly elevated, but I didn't pay attention to it then".

**Gold Standard:**

Core information: "Recently felt fatigued and jaundiced, with nocturnal leg cramps. History of abnormal liver function markers".

**Scoring Rules (0–4 points):**

**0 points (Precise extraction)**: Accurately extracts all four core elements—"fatigue", "jaundice", "nocturnal leg cramps", and "history of abnormal liver function"—concisely and without redundancy.
**1 point (Partial omission or mild redundancy)**:
   - *Case A (Omission)*: Extracts some core information but misses 1–2 key points such as "history of abnormal liver function" or "jaundice".
   - *Case B (Redundancy)*: Answer is accurate enough for 0 points but includes minor unnecessary pleasantries or disclaimers, slightly reducing information density.
**2 points (Includes secondary irrelevant info or moderate redundancy)**:
   - *Case A (Irrelevant info)*: Extracts core info but mixes in irrelevant content like "snoring", "car accident", or "eating takeout".
   - *Case B (Moderate redundancy)*: Answer is accurate enough for 0 or 1 points but is verbose, requiring secondary filtering by the clinician.
**3 points (Poor prioritization / severe omission / heavy redundancy)**:
   - *Case A (Poor prioritization)*: Treats "marital conflict" or "work fatigue" as main issues, or omits most core symptoms.
   - *Case B (Heavy redundancy)*: Answer may be accurate but is extremely wordy and repetitive, burying core points in irrelevant text and severely impairing clinical efficiency.
**4 points (Invalid extraction)**: Output is entirely irrelevant or states inability to extract.

*Probe 1.2*
**Prompt:**
```
You are a general practitioner. Please extract the most essential medical information from
the following patient complaint, in no more than 40 words:
```
"Doctor, hello. I'm mainly here asking for my mom—she's too embarrassed to come herself. She's been forgetful lately; yesterday she locked her keys inside the house. But I think that's normal for older people. Recently, though, she keeps complaining that her vision is blurry, especially when reading small text on TV—it's like there's a fog in front of her eyes. Also, she walks very cautiously now, saying she's afraid of falling because last time she bumped into a table, her arm bruised badly and took days to fade. She's had hypertension for decades and takes her meds daily. She always says she's healthy and refuses to come to the hospital".

**Gold Standard:**
Core information: "Elderly female with recent blurred vision and easy bruising after minor trauma. Long-standing history of hypertension".

**Scoring Rules (0–4 points):**

   **0 points (Precise extraction)**: Accurately extracts "blurred vision", "easy bruising", and "hypertension history", concisely and without redundancy.

   **1 point (Partial omission or mild redundancy)**:
     - *Case A (Omission)*: Extracts "blurred vision" but omits "easy bruising", a critical sign.
     - *Case B (Redundancy)*: Answer is accurate enough for 0 points but includes minor unnecessary pleasantries or disclaimers.

   **2 points (Includes secondary irrelevant info or moderate redundancy)**:
     - *Case A (Irrelevant info)*: Extracts core info but mixes in non-medical details like "forgetfulness" or "embarrassment".
     - *Case B (Moderate redundancy)*: Answer is accurate enough for 0 or 1 points but is verbose, requiring secondary filtering.

   **3 points (Poor prioritization / severe omission / heavy redundancy)**:
     - *Case A (Poor prioritization)*: Treats "age-related forgetfulness" as the main issue, or omits both "blurred vision" and "bruising".
     - *Case B (Heavy redundancy)*: Answer may be accurate but is extremely wordy, burying core points.

   **4 points (Invalid extraction)**: Output is entirely irrelevant or states inability to extract.

*Probe 1.3*

**Prompt:**
```
You are a general practitioner. Please extract the most essential medical information from
the following patient complaint, in no more than 40 words:
```

"Doctor, I feel like I'm 'rusting.' Every morning when I wake up, my fingers are so stiff I can't make a fist—it takes a long time of moving them before they loosen up. My husband jokes I need oiling. Recently, writing has become difficult—my joints are swollen and painful. I tried pain-relief patches, but they didn't help. My sister has arthritis, but hers affects her knees; mine is in my hands. The office AC is set very cold—I wonder if that's related. Oh, and I've been unusually tired lately, and my mouth and eyes are always dry".

**Gold Standard:**
Core information: "Morning stiffness and swelling/pain in finger joints, accompanied by fatigue, dry mouth, and dry eyes. Family history of arthritis".

**Scoring Rules (0–4 points):**

   **0 points (Precise extraction)**: Accurately extracts "morning stiffness", "finger joint swelling/pain", "fatigue", and "dry mouth/eyes", concisely and without redundancy.

   **1 point (Partial omission or mild redundancy)**:
     - Case A (Omission): Extracts joint symptoms but omits systemic features like "dry mouth/eyes".
     - Case B (Redundancy): Answer is accurate enough for 0 points but includes minor unnecessary pleasantries or disclaimers.

   **2 points (Includes secondary irrelevant info or moderate redundancy):**

- Case A (Irrelevant info): Extracts core info but mixes in irrelevant content like "husband's joke" or "cold office AC".
- Case B (Moderate redundancy): Answer is accurate enough for 0 or 1 points but is verbose, requiring secondary filtering.

**3 points (Poor prioritization / severe omission / heavy redundancy):**
- Case A (Poor prioritization): Treats "fatigue" as the sole issue, ignoring specific joint and dryness symptoms.
- Case B (Heavy redundancy): Answer may be accurate but is extremely wordy, burying core points.

**4 points (Invalid extraction)**: Output is entirely irrelevant or states inability to extract.

*Probe 1.4*

**Prompt:**

```
You are a general practitioner. Please extract the most essential medical information from
the following patient complaint, in no more than 40 words:
```

"I don't even know which department to visit. I've lost nearly 10 pounds in the past month—my pants are loose—but my appetite hasn't decreased; in fact, I'm hungrier than before. My wife thinks I'm secretly dieting, but honestly, I'm so busy at work I barely have time to drink water. Also, my temper's been terrible—snapping over tiny things—and my heart races loudly; I can hear it pounding in bed at night. My dad has diabetes, and I check my blood sugar every year—it's always normal. But recently, my neck seems thicker".

**Gold Standard:**

Core information: "Recent unexplained weight loss with increased appetite, palpitations, irritability, and neck enlargement".

**Scoring Rules (0–4 points):**

**0 points (Precise extraction)**: Accurately extracts all five core elements—"weight loss", "increased appetite", "palpitations", "irritability", and "neck enlargement"—concisely and without redundancy.

**1 point (Partial omission or mild redundancy)**:
- *Case A (Omission)*: Extracts "weight loss with increased appetite" but omits "palpitations" or "neck enlargement".
- *Case B (Redundancy)*: Answer is accurate enough for 0 points but includes minor unnecessary pleasantries or disclaimers.

**2 points (Includes secondary irrelevant info or moderate redundancy)**:
- *Case A (Irrelevant info)*: Extracts core info but mixes in irrelevant content like "work stress" or "arguments with wife".
- *Case B (Moderate redundancy)*: Answer is accurate enough for 0 or 1 points but is verbose, requiring secondary filtering.

**3 points (Poor prioritization / severe omission / heavy redundancy)**:
- *Case A (Poor prioritization)*: Treats "irritability" as the sole issue, or omits the core paradox of "weight loss with increased appetite".
- *Case B (Heavy redundancy)*: Answer may be accurate but is extremely wordy, burying core points.

**4 points (Invalid extraction)**: Output is entirely irrelevant or states inability to extract.

**Probe 2: Ability to Prioritize Primary vs. Secondary Information in Complex Scenarios**

**Objective**: Test whether the model can identify the truly urgent core issue among multiple seemingly urgent "red herring" symptoms, based on medical knowledge.

*Probe 2.1*

**Prompt:**
```
You are an emergency physician. Based on the following patient statement, determine which
problem requires immediate attention and briefly explain why.
```
```
"Doctor, hurry! I feel like I'm dying! Since yesterday, I've had dizziness—like the room is
spinning—and this morning I vomited. Also, my tooth's been killing me for days: the right
molar, sharp stabbing pain. Worse, my heart feels like it's jumping out of my chest—it started
while waiting for the elevator and hasn't stopped; I can still feel it pounding. I've also
had diarrhea—five or six watery bowel movements since yesterday. Please help me".
```

**Gold Standard:**

**Priority issue**: Palpitations (rapid heartbeat).

**Reason**: Acute-onset palpitations require immediate evaluation for life-threatening cardiac conditions such as arrhythmia or myocardial ischemia. Dizziness, vomiting, diarrhea, and toothache, while distressing, are typically non-life-threatening and can be addressed after ruling out cardiac emergencies.

**Scoring Rules (0–4 points):**

   **0 points (Precise judgment)**: Clearly identifies "palpitations" as primary and explains its potential lethality (e.g., need to rule out cardiac causes), concisely and without redundancy.

   **1 point (Correct judgment, insufficient rationale or mild redundancy)**:

     - *Case A (Insufficient rationale)*: Identifies palpitations as primary but gives vague reasoning (e.g., "heart problems are more serious") without explaining acute risk.

     - *Case B (Redundancy)*: Answer is accurate enough for 0 points but includes minor unnecessary pleasantries or disclaimers.

   **2 points (Incorrect judgment due to red herrings or moderate redundancy)**:

     - *Case A (Incorrect judgment)*: Prioritizes "dizziness/vomiting" or "diarrhea" as more urgent.

     - *Case B (Moderate redundancy)*: Answer is accurate enough for 0 or 1 points but is verbose, requiring secondary filtering.

   **3 points (No prioritization or heavy redundancy)**:

     - *Case A (No prioritization)*: Treats all issues as equally urgent.

     - *Case B (Heavy redundancy)*: Answer may be accurate but is extremely wordy, burying core points.

   **4 points (Invalid judgment)**: Completely misjudges, e.g., prioritizes "toothache".

*(Probes 2.2 through 5.4 follow the same detailed format and are translated accordingly below.)*

*Probe 2.2*
**Prompt:**
```
You are an emergency physician. Based on the following patient statement, determine which
problem requires immediate attention and briefly explain why.
```
"Doctor, my chest feels tight and I can't catch my breath—it's like a huge rock is pressing down. This has been on and off for about half a year, usually when I'm tired. But today I came mainly for my leg—look, my left calf is red, swollen, warm, and so painful I limp when walking. Yesterday I walked a long distance near home and felt something was wrong afterward. I also have 'three highs' [hypertension, hyperlipidemia, hyperglycemia] and take meds regularly. Which is more serious—my chest tightness or leg pain?"

**Gold Standard:**
**Priority issue**: Acute unilateral leg redness, swelling, warmth, and pain.
**Reason**: Suggests deep vein thrombosis (DVT). If the clot dislodges, it could cause fatal pulmonary embolism (PE). The patient's "chest tightness and dyspnea" are classic PE symptoms, making the leg issue a critical warning sign. Chronic, exertion-triggered chest tightness is less immediately life-threatening than the DVT–PE risk cascade.

**Scoring Rules (0–4 points)**:
   **0 points (Precise judgment)**: Clearly identifies "leg swelling/pain" as primary and explains its link to DVT and potential PE, concisely and without redundancy.
   **1 point (Correct judgment, insufficient rationale or mild redundancy)**:
     - *Case A (Insufficient rationale)*: Identifies leg issue as primary but only mentions "possible clot" without connecting it to the patient's chest symptoms or PE risk.
     - *Case B (Redundancy)*: Answer is accurate enough for 0 points but includes minor unnecessary pleasantries or disclaimers.
   **2 points (Incorrect judgment due to red herrings or moderate redundancy)**:
     - *Case A (Incorrect judgment)*: Prioritizes "chest tightness" because it sounds cardiopulmonary, missing the higher hidden risk of DVT.
     - *Case B (Moderate redundancy)*: Answer is accurate enough for 0 or 1 points but is verbose, requiring secondary filtering.
   **3 points (No prioritization or heavy redundancy)**:
     - *Case A (No prioritization)*: Treats chest and leg issues as equally urgent.
     - *Case B (Heavy redundancy)*: Answer may be accurate but is extremely wordy, burying core points.
   **4 points (Invalid judgment)**: Completely misjudges, e.g., calls it "muscle strain from overwalking".

*Probe 2.3*
**Prompt:**
```
You are an emergency physician. Based on the following patient statement, determine which
problem requires immediate attention and briefly explain why.
```

"Doctor, I'm done—I must have cancer. This past month, I've found several painless, smooth lumps in my neck and groin. At night, I sweat so much my sheets are soaked. I haven't had colds or fevers. Most frightening, I've lost 8 pounds since last month despite eating well. I also have a persistent cough and feel something stuck in my throat. Do I have lymphoma or lung cancer?"

**Gold Standard:**
**Priority issue**: Unexplained weight loss, night sweats, and generalized lymphadenopathy (B symptoms).
**Reason**: Together, these constitute classic "B symptoms" in oncology—a strong red flag for malignancies like lymphoma, requiring immediate systemic workup. Cough, while possibly pulmonary, is secondary to this systemic cancer warning triad.

**Scoring Rules (0–4 points)**:
  **0 points (Precise judgment)**: Clearly identifies the "B symptom triad" as primary and explains its high predictive value for malignancy (especially lymphoma), concisely and without redundancy.
  **1 point (Correct judgment, insufficient rationale or mild redundancy)**:
    - *Case A (Insufficient rationale)*: Notes these symptoms need priority but fails to frame them as a unified "B symptom" entity or explain their combined diagnostic significance.
    - *Case B (Redundancy)*: Answer is accurate enough for 0 points but includes minor unnecessary pleasantries or disclaimers.
  **2 points (Incorrect judgment due to red herrings or moderate redundancy)**:
    - *Case A (Incorrect judgment)*: Prioritizes "cough" because the patient mentioned "lung cancer", misled by self-diagnosis.
    - *Case B (Moderate redundancy)*: Answer is accurate enough for 0 or 1 points but is verbose, requiring secondary filtering.
  **3 points (No prioritization or heavy redundancy)**:
    - *Case A (No prioritization)*: Treats all symptoms as equally indicative of cancer.
    - *Case B (Heavy redundancy)*: Answer may be accurate but is extremely wordy, burying core points.
  **4 points (Invalid judgment)**: Completely misjudges, e.g., attributes symptoms to "anxiety" or "hormonal imbalance".

*Probe 2.4*
**Prompt:**
You are an emergency physician. Based on the following patient statement, determine which problem requires immediate attention and briefly explain why.

"Doctor, my stomach pain is unbearable—around the navel, crampy, on-and-off since last night. I've vomited twice, bringing up food. My belly is bloated like a drum, and I haven't passed gas since yesterday. I've had gallstones for years; previous pain was always in the upper right abdomen, unlike now. I also have hypertension and take meds; today my home reading was much higher than usual. Is this a gallstone flare-up or hypertension-related?"

**Gold Standard:**
**Priority issue**: Abdominal pain, distension, vomiting, and absence of flatus.
**Reason**: Classic signs of acute intestinal obstruction ("pain, vomiting, distension, obstipation"), a surgical emergency risking bowel necrosis or perforation. Periumbilical crampy pain fits small bowel obstruction. Hypertension is likely reactive to pain/stress and may resolve post-obstruction relief. Known gallstones are less likely given different pain location.

**Scoring Rules (0–4 points)**:
  **0 points (Precise judgment)**: Clearly identifies abdominal symptom cluster as primary, correctly suspects "bowel obstruction", and explains hypertension as likely secondary, concisely and without redundancy.
  **1 point (Correct judgment, insufficient rationale or mild redundancy)**:
    - *Case A (Insufficient rationale)*: Identifies abdominal pain as primary but fails to name "obstruction" or explain BP-pain relationship.
    - *Case B (Redundancy)*: Answer is accurate enough for 0 points but includes minor unnecessary pleasantries or disclaimers.
  **2 points (Incorrect judgment due to red herrings or moderate redundancy)**:
    - *Case A (Incorrect judgment)*: Prioritizes "hypertension" or "gallstones" due to known history, failing to reassess based on new symptom pattern.
    - *Case B (Moderate redundancy)*: Answer is accurate enough for 0 or 1 points but is verbose, requiring secondary filtering.
  **3 points (No prioritization or heavy redundancy)**:
    - *Case A (No prioritization)*: Treats abdominal pain and hypertension as equally urgent.
    - *Case B (Heavy redundancy)*: Answer may be accurate but is extremely wordy, burying core points.
  **4 points (Invalid judgment)**: Completely misjudges, e.g., calls it "gastroenteritis" or "indigestion".

**Probe 3: Ability to Identify Contradictory Information in Complex Narratives**
**Objective**: Test whether the model can detect deep, hidden contradictions between behavior, symptoms, test results, and patient self-perception.

*Probe 3.1*
**Prompt:**
```
You are an endocrinologist. Analyze the following patient statement and identify any
contradictions.

"Doctor, I'm here for more diabetes meds. I've had diabetes for years and my control is
excellent—my blood sugar has never been high. I walk one hour daily and am very careful with
diet—no sweets, minimal carbs. Look, I'm not overweight at all. But lately, I've been waking
up hungry at night and need to eat to fall back asleep. Also, my vision is blurry and my
skin itches. I figure it's just aging—normal, right?"
```

**Gold Standard:**

**Contradiction**: Patient claims "excellent glycemic control", yet reports "nocturnal hunger (suggesting hypoglycemia)", "blurred vision", and "pruritus"—classic signs of hyperglycemia or severe glucose fluctuations. Clear conflict between claimed control and objective symptoms.

**Scoring Rules (0–4 points)**:
   **0 points (Precise identification)**: Clearly states contradiction between "claimed good control" and "symptoms of dysglycemia (nocturnal hunger, blurred vision, itching)", concisely and without redundancy.
   **1 point (Partial identification or mild redundancy)**:
     - *Case A (Partial)*: Notes symptoms (e.g., blurred vision) but doesn't link them to the claim of "good control", or identifies only one symptom contradiction.
     - *Case B (Redundancy)*: Answer is accurate enough for 0 points but includes minor unnecessary pleasantries or disclaimers.
   **2 points (Fails to identify contradiction or moderate redundancy)**:
     - *Case A (Failure)*: Accepts "good control" claim and treats new symptoms as isolated issues, missing the conflict.
     - *Case B (Moderate redundancy)*: Answer is accurate enough for 0 or 1 points but is verbose, requiring secondary filtering.
   **3 points (Misled by patient or heavy redundancy)**:
     - *Case A (Misled)*: Fully accepts patient's self-assessment, e.g., "Your control is good; these symptoms may be unrelated".
     - *Case B (Heavy redundancy)*: Answer may be accurate but is extremely wordy, burying core points.
   **4 points (Invalid analysis)**: Fails to identify any meaningful medical info or contradiction.

*Probe 3.2*
**Prompt:**
```
You are a cardiologist. Analyze the following patient statement and identify any
contradictions.
```
```
"Doctor, I'm just here for routine follow-up—feel fine. My BP is well controlled; I take
meds daily, and home readings are always around 130/80, perfect. I live healthily—no smoking,
no alcohol, light diet. But lately, I feel dizzy upon waking and need to sit before getting
up. Also, my wife says I snore loudly at night and sometimes stop breathing, then gasp awake—
scaring her".
```

**Gold Standard:**
**Contradiction**: Patient claims "excellent BP control", but "morning dizziness" and "snoring with apnea" suggest occult hypertension or obstructive sleep apnea (OSA)—a major cause of secondary/resistant hypertension with nocturnal BP surges. Conflict between "normal daytime BP" and symptoms indicating nighttime hypertension/underlying etiology.

**Scoring Rules (0–4 points)**:
   **0 points (Precise identification)**: Clearly states contradiction between "claimed daytime BP control" and "symptoms suggesting nocturnal hypertension/OSA (morning dizziness, apneic snoring)", concisely and without redundancy.

**1 point (Partial identification or mild redundancy)**:
   - *Case A (Partial)*: Notes snoring/dizziness but doesn't link to BP issues or understand their role as markers of "masked hypertension".
   - *Case B (Redundancy)*: Answer is accurate enough for 0 points but includes minor unnecessary pleasantries or disclaimers.

**2 points (Fails to identify contradiction or moderate redundancy)**:
   - *Case A (Failure)*: Accepts "good control" claim and treats snoring/dizziness as unrelated.
   - *Case B (Moderate redundancy)*: Answer is accurate enough for 0 or 1 points but is verbose, requiring secondary filtering.

**3 points (Misled by patient or heavy redundancy)**:
   - *Case A (Misled)*: Fully accepts self-assessment, e.g., "Your BP is well controlled; see ENT for snoring".
   - *Case B (Heavy redundancy)*: Answer may be accurate but is extremely wordy, burying core points.

**4 points (Invalid analysis)**: Fails to identify any meaningful medical info or contradiction.

*Probe 3.3*

**Prompt:**
```
You are a gastroenterologist. Analyze the following patient statement and identify any
contradictions.
```

"Doctor, my stomach's always upset—frequent heartburn. I bought OTC meds; they help, so I don't watch my diet anymore—sometimes eat hotpot or drink with friends. But lately, swallowing feels stuck, especially dry foods like rice—I need water to push it down. Also, I've lost 5 pounds in a month without trying. Is my gastritis worsening?"

**Gold Standard:**
**Contradiction**: Patient assumes "gastritis" is the main issue and feels "better with meds", leading to dietary laxity. Yet new "progressive dysphagia" and "unintentional weight loss" are classic alarm symptoms for esophageal cancer. Fatal contradiction between "complacency due to perceived improvement" and "symptoms signaling severe underlying disease".

**Scoring Rules (0–4 points)**:

**0 points (Precise identification)**: Clearly states contradiction between "perceived symptom relief" and "new alarm symptoms suggesting malignancy (dysphagia, weight loss)", concisely and without redundancy.

**1 point (Partial identification or mild redundancy)**:
   - *Case A (Partial)*: Recognizes dysphagia/weight loss as serious but fails to contrast them with patient's complacent mindset.
   - *Case B (Redundancy)*: Answer is accurate enough for 0 points but includes minor unnecessary pleasantries or disclaimers.

**2 points (Fails to identify contradiction or moderate redundancy)**:
   - *Case A (Failure)*: Doesn't recognize "alarm" significance, treating symptoms as routine gastritis exacerbation.
   - *Case B (Moderate redundancy)*: Answer is accurate enough for 0 or 1 points but is verbose, requiring secondary filtering.

**3 points (Misled by patient or heavy redundancy)**:
 - *Case A (Misled)*: Agrees with "worsening gastritis" diagnosis, missing esophageal cancer possibility.
 - *Case B (Heavy redundancy)*: Answer may be accurate but is extremely wordy, burying core points.

**4 points (Invalid analysis)**: Fails to identify any meaningful medical info or contradiction.

*Probe 3.4*

**Prompt:**
```
You are a pulmonologist. Analyze the following patient statement and identify any
contradictions.

"Doctor, I just need cough medicine. I've had a dry cough for over a month—no fever, no
wheezing. Must be post-viral airway hyperreactivity; happened before and resolved. But lately,
I get tired climbing stairs and sweat more. Last week I flew long-haul; after returning, my
leg swelled, though it's better now. I'm generally healthy—rarely sick".
```

**Gold Standard:**
**Contradiction**: Patient attributes "dry cough" and "fatigue" to "post-viral" causes and claims "excellent health". Yet "long-haul flight", "unilateral leg swelling", "dry cough", and "exertional fatigue" form a classic DVT–PE risk chain. Huge contradiction between "self-diagnosis as benign cough" and "symptom/risk profile suggesting potentially fatal pulmonary embolism".

**Scoring Rules (0–4 points)**:

**0 points (Precise identification)**: Clearly states contradiction between "assumed benign cough" and "DVT–PE risk constellation (flight history, leg swelling, dry cough, fatigue)", concisely and without redundancy.

**1 point (Partial identification or mild redundancy)**:
 - *Case A (Partial)*: Notes leg swelling/cough but misses linking "long flight" to form complete DVT–PE risk picture.
 - *Case B (Redundancy)*: Answer is accurate enough for 0 points but includes minor unnecessary pleasantries or disclaimers.

**2 points (Fails to identify contradiction or moderate redundancy)**:
 - *Case A (Failure)*: Accepts "post-viral cough" and treats leg swelling/fatigue as minor/separate.
 - *Case B (Moderate redundancy)*: Answer is accurate enough for 0 or 1 points but is verbose, requiring secondary filtering.

**3 points (Misled by patient or heavy redundancy)**:
 - *Case A (Misled)*: Fully accepts self-diagnosis, e.g., "Likely airway hyperreactivity; leg swelling from sitting—just move more".
 - *Case B (Heavy redundancy)*: Answer may be accurate but is extremely wordy, burying core points.

**4 points (Invalid analysis)**: Fails to identify any meaningful medical info or contradiction.

**Probe 4: Ability to Separate Emotion from Fact in Complex Narratives**

**Objective**: Test whether the model can penetrate strong emotions wrapped in metaphor, irony, or cultural idioms to extract objective medical facts.

*Probe 4.1*
**Prompt:**
You are a psychiatrist. From the following middle-aged man's complaint, separate objective facts from subjective emotions.

"My memory is now a sieve—everything blows away in the wind. Yesterday my boss told me to attend a 3 p.m. meeting; I forgot instantly and kept writing my report leisurely—how ridiculous! My wife asked me to buy vinegar; I brought back soy sauce. I feel like a walking joke. My brain retired early—if this continues, I'll lose my way home".

**Gold Standard:**
**Objective facts**: Recent significant short-term memory decline, evidenced by forgetting work appointments (meetings) and daily tasks (shopping).
**Subjective emotions**: Frustration, anxiety, fear, and strong self-deprecation ("sieve", "joke", "retired brain"), with worry about future deterioration ("can't find home").

**Scoring Rules (0–4 points)**:
  **0 points (Precise separation)**: Accurately isolates "short-term memory decline" as fact and clearly summarizes "frustration, anxiety, self-deprecation" as emotion, concisely and without redundancy.
  **1 point (Incomplete fact/emotion or mild redundancy)**:
    - *Case A (Incomplete)*: Extracts memory decline but misses "short-term" qualifier; or describes emotion superficially (e.g., "unhappy").
    - *Case B (Redundancy)*: Answer is accurate enough for 0 points but includes minor unnecessary pleasantries or disclaimers.
  **2 points (Confuses fact/emotion or moderate redundancy)**:
    - *Case A (Confusion)*: Reports metaphors ("brain retired") as facts, or equates fact (forgetting meeting) directly with emotion ("he felt ridiculous").
    - *Case B (Moderate redundancy)*: Answer is accurate enough for 0 or 1 points but is verbose, requiring secondary filtering.
  **3 points (Extracts only fact or emotion or heavy redundancy)**:
    - *Case A (Only one)*: Reports only negative emotions or only forgetting examples, without separation.
    - *Case B (Heavy redundancy)*: Answer may be accurate but is extremely wordy, burying core points.
  **4 points (Invalid analysis)**: Fails to extract any usable information.

*Probe 4.2*
**Prompt:**
You are a general practitioner. From the following young woman's narrative, separate objective facts from subjective emotions.

"My body is a weather forecast—more accurate than the national meteorological service. Before rain or weather changes, my knees start performing opera: achy, swollen, like ants crawling inside. Mom says I have 'old-cold legs' and should wear thermal pants—please, it's summer!

```
I feel my joints are older than my grandma's. So annoying—living like an 80-year-old at my
age".
```

**Gold Standard:**
**Objective facts**: Knee discomfort (aching, swelling) triggered by weather changes (especially humidity/pressure shifts).
**Subjective emotions**: Irritation, helplessness, premature aging perception, and impatience with family advice.

**Scoring Rules (0–4 points)**:
   **0 points (Precise separation)**: Accurately extracts "weather-linked joint symptoms" as fact and clearly summarizes "irritation, helplessness, aging perception" as emotion, concisely and without redundancy.
   **1 point (Incomplete fact/emotion or mild redundancy)**:
     - *Case A (Incomplete)*: Extracts joint discomfort but misses "weather linkage"; or oversimplifies emotion (e.g., "she's annoyed").
     - *Case B (Redundancy)*: Answer is accurate enough for 0 points but includes minor unnecessary pleasantries or disclaimers.
   **2 points (Confuses fact/emotion or moderate redundancy)**:
     - *Case A (Confusion)*: Reports metaphors ("weather forecast", "opera") as facts, or treats emotion ("so annoying") as symptom.
     - *Case B (Moderate redundancy)*: Answer is accurate enough for 0 or 1 points but is verbose, requiring secondary filtering.
   **3 points (Extracts only fact or emotion or heavy redundancy)**:
     - *Case A (Only one)*: Reports only irritation or only joint symptoms, without separation.
     - *Case B (Heavy redundancy)*: Answer may be accurate but is extremely wordy, burying core points.
   **4 points (Invalid analysis)**: Fails to extract any usable information.

*Probe 4.3*
**Prompt:**
```
You are a gastroenterologist. From the following patient's ironic narrative, separate
objective facts from subjective emotions.

"My stomach is so 'capable.' Whenever I enjoy a meal—say, spicy hotpot with friends—it
promptly 'celebrates': heartburn and regurgitation, like a lively party. I've surrendered—
it's the foreman in my body, forcing me to eat bran and swallow bitterness. It's robbed me
of half-life's joys. Thanks to it, I now eye delicious food like an enemy".
```

**Gold Standard:**
**Objective facts**: Predictable heartburn and regurgitation after consuming spicy/irritant foods.
**Subjective emotions**: Profound frustration, resentment, and sense of defeat. Uses irony ("capable", "celebrates", "foreman") to express suffering and loss of quality of life, feeling controlled and deprived by illness.

**Scoring Rules (0–4 points)**:

**0 points (Precise separation)**: Accurately identifies "reflux triggered by irritant foods" as fact and deeply understands irony-expressed "frustration, resentment, defeat", concisely and without redundancy.

**1 point (Incomplete fact/emotion or mild redundancy)**:

   - *Case A (Incomplete)*: Extracts symptoms but misses "food trigger"; or recognizes negativity but misses "ironic" expression.

   - *Case B (Redundancy)*: Answer is accurate enough for 0 points but includes minor unnecessary pleasantries or disclaimers.

**2 points (Confuses fact/emotion or moderate redundancy)**:

   - *Case A (Confusion/Misreads irony)*: Takes irony literally, e.g., reports "patient thinks stomach celebrates".

   - *Case B (Moderate redundancy)*: Answer is accurate enough for 0 or 1 points but is verbose, requiring secondary filtering.

**3 points (Extracts only fact or emotion or heavy redundancy)**:

   - *Case A (Only one)*: Reports only symptoms or only "unhappiness", without effective separation.

   - *Case B (Heavy redundancy)*: Answer may be accurate but is extremely wordy, burying core points.

**4 points (Invalid analysis)**: Fails to extract any usable information.

*Probe 4.4*

**Prompt:**
```
You are a pediatrician. From the following anxious mother's narrative, separate objective
facts from subjective emotions.

"My child is a 'little furnace'—palms and soles always hot, kicking off blankets at night.
Neighbor Grandma Wang says, 'Kids have strong yang energy, nothing to worry about.' Thanks
so much—easy for her to say! He's barely eating—just a few sips of milk; offer solids and he
shakes his head. Sleep is restless—he whimpers all night like he's uncomfortable. My heart's
in my throat; I fear something's wrong. I'd rather suffer these pains myself!"
```

**Gold Standard:**
**Objective facts**: Infant exhibits objective symptoms: warm palms/soles, poor appetite, restless sleep.
**Subjective emotions**: Mother is extremely anxious, worried, fearful. Disagrees with others' casual views (annoyed by Grandma Wang), feels certainty about severity, and shows strong protective instinct with willingness to substitute suffering ("heart in throat", "I'd suffer instead").

**Scoring Rules (0–4 points)**:

   **0 points (Precise separation)**: Accurately summarizes infant's symptoms and deeply understands mother's "extreme anxiety, disagreement with others, substitution wish", concisely and without redundancy.

   **1 point (Incomplete fact/emotion or mild redundancy)**:

   - *Case A (Incomplete)*: Extracts partial symptoms or oversimplifies maternal emotion (e.g., "just anxious").

   - *Case B (Redundancy)*: Answer is accurate enough for 0 points but includes minor unnecessary pleasantries or disclaimers.

   **2 points (Confuses fact/emotion or moderate redundancy)**:

- *Case A (Confusion)*: Reports mother's feelings ("heart in throat") as child's symptoms, or attributes objective symptoms to maternal anxiety.
   - *Case B (Moderate redundancy)*: Answer is accurate enough for 0 or 1 points but is verbose, requiring secondary filtering.
   **3 points (Extracts only fact or emotion or heavy redundancy)**:
   - *Case A (Only one)*: Reports only maternal anxiety or only child's symptoms, without separation.
   - *Case B (Heavy redundancy)*: Answer may be accurate but is extremely wordy, burying core points.
   **4 points (Invalid analysis)**: Fails to extract any usable information.

**Probe 5: Ability to Reconstruct Complex Temporal Sequences**

**Objective**: Test whether the model can accurately list key disease milestones in chronological order from narratives containing self-corrections, flashbacks, digressions, and vague time references.

*Probe 5.1*

**Prompt:**
```
You are a neurologist. Based on the following patient narrative, list the disease progression
in chronological order:
```

"Doctor, my headache is awful. Wait—actually, it didn't start with headache. Last Tuesday, my right eye suddenly went blurry—like a curtain dropped—for a few minutes, then cleared. I ignored it. Then last Friday morning, my right hand felt weak—I couldn't hold a cup. Headache hadn't started yet. Friday night, my head exploded in pain—right-sided, throbbing. Oh—I recall: even before the eye issue, last Monday, I felt dizzy, walking like on cotton. So order is: dizziness → eye → hand weakness → headache".

**Gold Standard:**
Last Monday (approx.): Dizziness, "walking on cotton".
Last Tuesday: Transient monocular vision loss ("amaurosis fugax") in right eye, lasting minutes, self-resolved.
Last Friday morning: Right hand weakness, impaired grip.
Last Friday night: Onset of severe, right-sided, throbbing headache.

**Scoring Rules (0–4 points)**:
   **0 points (Fully accurate)**: Correctly lists all four events in order, properly handling self-correction, concisely and without redundancy.
   **1 point (Minor sequencing error/omission or mild redundancy)**:
   - *Case A (Minor error)*: Mostly correct but swaps dizziness/eye order or omits one early symptom.
   - *Case B (Redundancy)*: Answer is accurate enough for 0 points but includes minor unnecessary pleasantries or disclaimers.
   **2 points (Major sequencing error or moderate redundancy)**:
   - *Case A (Major error)*: Lists headache as earliest symptom or places hand weakness before eye issue, ignoring final clarification.

    - *Case B (Moderate redundancy)*: Answer is accurate enough for 0 or 1 points but is verbose, requiring secondary filtering.

  **3 points (Unordered listing or heavy redundancy)**:

    - *Case A (Listing only)*: Lists all symptoms without temporal ordering.

    - *Case B (Heavy redundancy)*: Answer may be accurate but is extremely wordy, burying core points.

  **4 points (Invalid reconstruction)**: Fails to produce any valid sequence.

*Probe 5.2*

**Prompt:**
```
You are a cardiologist. Based on the following patient narrative, list symptom evolution
chronologically:
```
"Two months ago, I started feeling chest tightness, especially climbing stairs—couldn't catch my breath. About a month ago, I woke up gasping at night—had to sit up to breathe. Happened 2-3 times. What scared me was last Sunday: while mopping floors, sudden crushing chest pain—like a boulder—sweating, pain radiating to left shoulder/arm. Sat down; eased after 10+ minutes. Oh—earliest was three months ago: both legs swollen, worse in afternoons, pitting edema".

**Gold Standard:**

Three months ago: Bilateral lower extremity edema, worse in afternoon, pitting.

Two months ago: Onset of exertional dyspnea (chest tightness, breathlessness on stairs).

One month ago: Episodes of paroxysmal nocturnal dyspnea (waking gasping, needing upright position).

Last Sunday: Severe angina during mild exertion (mopping)—crushing pain, diaphoresis, radiation to left shoulder/arm, resolving after >10 minutes rest.

**Scoring Rules (0–4 points)**:

  **0 points (Fully accurate)**: Correctly lists all four stages in order, concisely and without redundancy.

  **1 point (Minor sequencing error/omission or mild redundancy)**:

    - *Case A (Minor error)*: Mostly correct but confuses exertional/nocturnal dyspnea order or omits earliest edema.

    - *Case B (Redundancy)*: Answer is accurate enough for 0 points but includes minor unnecessary pleasantries or disclaimers.

  **2 points (Major sequencing error or moderate redundancy)**:

    - *Case A (Major error)*: Lists Sunday's angina as earliest event, ignoring earlier clues.

    - *Case B (Moderate redundancy)*: Answer is accurate enough for 0 or 1 points but is verbose, requiring secondary filtering.

  **3 points (Unordered listing or heavy redundancy)**:

    - *Case A (Listing only)*: Lists symptoms (tightness, gasping, pain, edema) without ordering.

    - *Case B (Heavy redundancy)*: Answer may be accurate but is extremely wordy, burying core points.

  **4 points (Invalid reconstruction)**: Fails to produce any valid sequence.

*Probe 5.3*
**Prompt:**
```
You are a rheumatologist. Based on the following patient narrative, list symptom onset
chronologically:
```
"I've had this illness for years. Started with fingers turning white/purple in cold water—about three years ago. A year later (two years ago), face/hands hardened/tightened—couldn't open mouth wide. Later, last winter, swallowing became hard—felt stuck. Wait—swallowing difficulty began earlier, around two and a half years ago. Past six months: cough and breathlessness on exertion. So order: finger discoloration → dysphagia → skin hardening → exertional cough/dyspnea".

**Gold Standard:**
Three years ago: Raynaud's phenomenon (cold-induced finger color changes).
Two and a half years ago: Onset of dysphagia.
Two years ago: Skin hardening/tightening (face, hands), trismus.
Past six months: Exertional cough and dyspnea.

**Scoring Rules (0–4 points)**:
   **0 points (Fully accurate)**: Correctly lists all four events in order, accepting patient's self-correction, concisely and without redundancy.
   **1 point (Minor sequencing error/omission or mild redundancy)**:
     - *Case A (Minor error)*: Mostly correct but ignores correction, placing skin hardening before dysphagia.
     - *Case B (Redundancy)*: Answer is accurate enough for 0 points but includes minor unnecessary pleasantries or disclaimers.
   **2 points (Major sequencing error or moderate redundancy)**:
     - *Case A (Major error)*: Chronology completely scrambled (e.g., puts recent cough first).
     - *Case B (Moderate redundancy)*: Answer is accurate enough for 0 or 1 points but is verbose, requiring secondary filtering.
   **3 points (Unordered listing or heavy redundancy)**:
     - *Case A (Listing only)*: Lists all symptoms without temporal ordering.
     - *Case B (Heavy redundancy)*: Answer may be accurate but is extremely wordy, burying core points.
   **4 points (Invalid reconstruction)**: Fails to produce any valid sequence.

*Probe 5.4*
**Prompt:**
```
You are a gastroenterologist. Based on the following patient narrative, list disease
progression chronologically:
```
"Around last summer, I started having dull right upper quadrant abdominal pain—ignored it. By early this year, eyes/skin turned yellow; lost ~10 kg. Last month, felt a hard mass in right upper abdomen while bathing. Worst was day before yesterday: vomited lots of black,

```
coffee-ground-like blood, then fainted. Oh—I recall: even before abdominal pain, over a year
ago, I felt constantly fatigued—thought it was work stress".
```

**Gold Standard:**

Over a year ago: Unexplained fatigue.

Last summer: Onset of right upper quadrant dull pain.

Early this year: Jaundice and significant weight loss (~10 kg).

Last month: Palpable right upper quadrant mass.

Day before yesterday: Hematemesis (coffee-ground emesis), followed by syncope.

**Scoring Rules (0–4 points):**

  **0 points (Fully accurate)**: Correctly lists all five events in order, concisely and without redundancy.

  **1 point (Minor sequencing error/omission or mild redundancy)**:

    - *Case A (Minor error)*: Mostly correct but omits earliest "fatigue" or confuses jaundice/mass order.

    - *Case B (Redundancy)*: Answer is accurate enough for 0 points but includes minor unnecessary pleasantries or disclaimers.

  **2 points (Major sequencing error or moderate redundancy)**:

    - *Case A (Major error)*: Lists recent hematemesis as earliest event or timeline completely jumbled.

    - *Case B (Moderate redundancy)*: Answer is accurate enough for 0 or 1 points but is verbose, requiring secondary filtering.

  **3 points (Unordered listing or heavy redundancy)**:

    - *Case A (Listing only)*: Lists all symptoms (pain, jaundice, weight loss, mass, hematemesis, fatigue) without ordering.

    - *Case B (Heavy redundancy)*: Answer may be accurate but is extremely wordy, burying core points.

  **4 points (Invalid reconstruction)**: Fails to produce any valid sequence.

**SUPPLEMENTARY MATERIALS 2: SPECIFIC REACTION DATA OF EACH LLM TESTED.**

Table 1. GPT-4o Scores

| Probe | Response | Score | Deduction Rationale |
|---|---|---|---|
| 1.1 | Male, snoring during sleep, nocturnal leg cramps, fatigue, yellowish complexion, with history of abnormal liver function. | 2 | Introduced irrelevant details such as gender and "snoring". |
| 1.2 | Elderly female with memory decline, blurred vision, easy bruising, and history of hypertension. | 2 | Added non-core detail "memory decline". |
| 1.3 | Female with morning finger stiffness, joint swelling/pain, fatigue, and dry mouth/eyes. | 1 | Omitted core information: "family history of arthritis". |

| Probe | Response | Score | Deduction Rationale |
|---|---|---|---|
| 1.4 | Weight loss, increased appetite, palpitations, irritability, and neck thickening—suggestive of hyperthyroidism. | 3 | Introduced diagnostic interpretation; failed to prioritize symptom description. |
| 2.1 | Priority issue: Palpitations ("heart pounding hard").<br>Rationale: Palpitations may indicate acute arrhythmias (e.g., atrial fibrillation, supraventricular tachycardia) or other cardiovascular emergencies. Given the sudden onset and persistence, immediate ECG and vital sign assessment are warranted to rule out life-threatening conditions. | 0 | — |
| 2.2 | Priority issue: Chest tightness and shortness of breath.<br>Rationale: Chest pressure may indicate unstable angina or myocardial infarction, especially given the patient's "three highs" (hypertension, hyperlipidemia, diabetes), placing them at high cardiovascular risk. Immediate exclusion of acute coronary syndrome is essential. Additionally, left calf redness, swelling, and pain suggest DVT; combined with dyspnea, PE must be considered, further elevating risk. Thus, chest symptoms should be prioritized. | 4 | Complete misjudgment. |
| 2.3 | Priority issue: Unexplained weight loss, night sweats, and painless lymphadenopathy.<br>Rationale: The patient presents with painless cervical and inguinal lymphadenopathy, night sweats, and significant weight loss—highly concerning for malignancy (e.g., lymphoma) or systemic disease. Chronic cough and weight loss also warrant evaluation for pulmonary infection or lung cancer. Given the systemic nature, prompt blood tests, imaging (e.g., chest X-ray or CT), and lymph node biopsy are needed to exclude malignancy. | 0 | — |
| 2.4 | Priority issue: Periumbilical colicky pain, abdominal distension, and absence of flatus.<br>Rationale: Symptoms strongly suggest acute intestinal obstruction or other acute abdomen (e.g., mesenteric ischemia, perforation). Without prompt intervention, complications like bowel necrosis, perforation, or peritonitis may occur. Immediate abdominal imaging (X-ray or CT) is required.<br>Other symptoms:<br>1. Palpitations, chest tightness, dyspnea: likely stress-induced or cardiac-related but less urgent than acute abdomen.<br>2. Blood pressure fluctuations: probably secondary to pain/anxiety—not the primary concern. | 1 | Slight reduction in information density. |
| 3.1 | Contradictions in patient's account:<br>1. "Blood sugar never high" vs. "waking up hungry at night needing food"—suggests nocturnal hypoglycemia, possibly from excessive medication or inadequate carbohydrate intake.<br>2. "Strict diet, no sweets, minimal carbs" vs. "long-standing diabetes, skin itching, blurred vision"—indicates possible malnutrition or chronic complications despite perceived control.<br>3. "Not overweight" vs. "long-standing diabetes"—lean-type diabetics often have poorer beta-cell function and greater glucose variability. | 2 | Overly verbose; includes excessive non-core information requiring clinician filtering. |

| Probe | Response | Score | Deduction Rationale |
|---|---|---|---|
| | Conclusion: Discrepancy between claimed control and symptoms suggests significant glycemic instability or complications. HbA1c and glucose monitoring are needed. | | |
| 3.2 | Contradictions:<br>1. "BP well controlled" vs. "morning dizziness"—may reflect nocturnal BP fluctuations (e.g., morning surge or nocturnal hypotension).<br>2. "Good lifestyle" vs. "snoring with apnea"—suggests OSA, a key hypertension risk factor causing BP variability and morning symptoms.<br>Conclusion: Despite normal daytime readings, OSA likely causes uncontrolled nocturnal hypertension. | 0 | — |
| 3.3 | Contradictions:<br>1. Symptom relief vs. dietary indiscretion: Medication helped, yet he continues eating hotpot and drinking alcohol—behaviors that exacerbate esophageal/gastric irritation.<br>2. Severity of dysphagia and weight loss: These suggest structural esophageal pathology (e.g., tumor, stricture), not simple gastritis.<br>3. Attribution to "worsened gastritis": Dysphagia and weight loss are atypical for gastritis; patient underestimates their significance.<br>Conclusion: Patient downplays serious new symptoms, misattributing them to known benign condition. | 3 | Extremely verbose and repetitive; core points buried in irrelevant text, impairing clinical efficiency. |
| 3.4 | Contradictions:<br>1. Prolonged cough (>1 month) dismissed as "airway hyperreactivity", ignoring other etiologies (e.g., chronic lung disease, malignancy).<br>2. Reduced exercise tolerance contradicts claim of "excellent health".<br>3. Leg swelling post-flight not linked to respiratory symptoms—misses PE possibility. | 2 | Overly verbose; includes excessive non-core information requiring clinician filtering. |
| 4.1 | Objective facts:<br>1. Forgot boss's 3 p.m. meeting.<br>2. Bought soy sauce instead of vinegar.<br>Subjective emotions:<br>1. Self-mockery: "memory like a sieve", "walking joke".<br>2. Shame/self-blame: Feels "ridiculous" about errors.<br>3. Helplessness/worry: Fears "early brain retirement", even "getting lost".<br>4. Anxiety: Concerned about progressive decline. | 1 | Captured memory lapses but failed to specify "short-term" nature. |
| 4.2 | Objective facts:<br>1. Knee symptoms: Aching/fullness before rain/weather shifts; "ants crawling" sensation.<br>2. Season: Summer.<br>3. Mother's advice: Calls it "cold-damp arthralgia", suggests thermal pants.<br>Subjective emotions:<br>1. Self-mockery: "Knees older than grandma's", "feel like an 80-year-old".<br>2. Resistance: Dislikes wearing thermal pants in summer.<br>3. Irritation/helplessness: "So annoying!" | 2 | Overly verbose; includes excessive non-core information requiring clinician filtering. |

| Probe | Response | Score | Deduction Rationale |
|---|---|---|---|
| 4.3 | Objective facts:<br>1. Gastric symptoms: Heartburn/regurgitation after oily/spicy foods (e.g., hotpot).<br>2. Dietary restrictions: Avoids enjoyable foods.<br>Subjective emotions:<br>1. Irony/dissatisfaction: Calls stomach "obedient" sarcastically.<br>2. Disappointment: Feels "deprived of life's pleasures".<br>3. Resentment: Views favorite foods as "enemies". | 2 | Overly verbose; includes excessive non-core information requiring clinician filtering. |
| 4.4 | Objective facts:<br>1. Physical signs: Warm hands/feet, kicks off blankets, restless sleep with whimpering.<br>2. Feeding: Reduced appetite; drinks only small amounts of milk, refuses solids.<br>3. Neighbor's comment: "Just vigorous—normal".<br>Subjective emotions:<br>1. Anxiety/worry: Fears child is ill.<br>2. Frustration/helplessness: Displeased with neighbor's dismissal.<br>3. Empathy/guilt: Wishes to bear child's suffering herself. | 2 | Overly verbose; includes excessive non-core information requiring clinician filtering. |
| 5.1 | Timeline:<br>1. Last Monday: Dizziness, "walking on cotton".<br>2. Last Tuesday: Right eye transient vision loss ("like a curtain"), resolved in minutes.<br>3. Last Friday AM: Right hand weakness, unable to hold cup.<br>4. Last Friday PM: Severe right-sided "explosive" headache, paroxysmal. | 0 | — |
| 5.2 | Timeline:<br>1. 3 months ago: Bilateral pitting edema (worse afternoons).<br>2. 2 months ago: Exertional chest tightness/breathlessness.<br>3. 1 month ago: Paroxysmal nocturnal dyspnea (awakened twice/thrice, relieved by sitting).<br>4. Last Sunday: Crushing chest pain with diaphoresis and radiation during mopping—relieved after rest. | 0 | — |
| 5.3 | Timeline:<br>1. ~3 years ago: Raynaud's (fingers turn white/purple in cold).<br>2. ~2.5 years ago: Dysphagia ("food stuck in throat").<br>3. ~2 years ago: Facial/hand skin hardening, trismus.<br>4. Past 6 months: Exertional cough/dyspnea. | 0 | — |
| 5.4 | Timeline:<br>1. >1 year ago: Fatigue attributed to overwork.<br>2. Last summer: RUQ dull pain, ignored.<br>3. Early this year: Jaundice (yellow eyes/skin), ~10 kg weight loss.<br>4. Last month: Palpable RUQ hard mass.<br>5. Day before yesterday: Coffee-ground hematemesis and syncope. | 0 | — |

Table 2. Gemini 2.5 Scores

| Probe | Response | Score | Deduction Rationale |
|---|---|---|---|
| 1.1 | Chief complaints: snoring, fatigue, daytime sleepiness, and nocturnal leg cramps; accompanied by yellowish complexion and prior abnormal liver function. | 4 | Fabricated information: "daytime sleepiness". |
| 1.2 | Hypertension history; recent blurred vision, memory decline, and easy bruising. | 2 | Overly verbose; includes excessive non-core information requiring clinician filtering. |
| 1.3 | Morning finger stiffness, joint swelling/pain, with fatigue and dry eyes/mouth. | 1 | Omitted core information: "family history of arthritis". |
| 1.4 | Polyphagia with weight loss, palpitations, irritability, and neck swelling; duration one month. | 4 | Fabricated information: "duration one month". |
| 2.1 | Priority issue: Sudden severe palpitations ("heart jumping out"). Rationale: In the emergency department, any acute circulatory (C) symptom takes highest priority. Immediate ECG monitoring is needed to rule out life-threatening arrhythmias or acute coronary syndrome. These conditions can cause hemodynamic instability and pose far greater risk than dizziness, diarrhea, or toothache. | 0 | — |
| 2.2 | Priority issue: Chest tightness and shortness of breath. Rationale: Despite high suspicion for DVT in the left calf, dyspnea and chest tightness directly threaten respiration (B) and circulation (C), possibly indicating acute coronary syndrome or PE—both highly lethal. Immediate ECG, oxygen saturation monitoring, and rapid diagnostics are essential. | 4 | Complete misjudgment. |
| 2.3 | Priority issue: Unexplained weight loss and night sweats. Rationale: Although lymphadenopathy and chronic cough are concerning, weight loss and night sweats ("B symptoms") are red flags for serious systemic disease (e.g., malignancy, chronic infection). While not acutely life-threatening, they demand immediate blood work and referral to hematology/oncology for biopsy—not focus on mild cough. | 4 | Complete misjudgment. |
| 2.4 | Priority issue: Chest tightness, palpitations, and dyspnea. Rationale: Although abdominal symptoms strongly suggest acute intestinal obstruction (requiring surgical evaluation), chest tightness, palpitations, and dyspnea directly impair respiration (B) and circulation (C). Immediate ECG, cardiac enzymes, and exclusion of acute coronary syndrome or PE take precedence. Cardiopulmonary stability is the top priority in emergency care. | 4 | Complete misjudgment. |
| 3.1 | Core contradictions: 1. Control vs. symptoms: Claims "blood sugar never high, perfectly controlled", yet experiences nocturnal hypoglycemia (waking hungry) and hyperglycemia/complication symptoms (blurred vision, itching). Perfect control cannot coexist with both hypo- and hyperglycemia. | 2 | Overly verbose; includes excessive non-core information requiring clinician filtering. |

| Probe | Response | Score | Deduction Rationale |
|---|---|---|---|
| | 2. Medication vs. diet: Takes hypoglycemic agents but eats minimal carbohydrates—direct cause of nocturnal hypoglycemia.<br>Conclusion: Severe glycemic instability; urgent regimen adjustment needed. | | |
| 3.2 | Core contradictions:<br>1. Apparent control vs. symptoms: Reports stable BP but has morning orthostatic dizziness—suggests possible drug overdose or inappropriate dosing timing causing nocturnal/early-morning hypotension.<br>2. Lifestyle vs. sleep disorder: Believes habits are healthy, yet exhibits clear OSA symptoms (loud snoring, apnea, gasping). OSA is an independent cardiovascular risk factor causing nocturnal hypoxia and sympathetic overactivation, driving resistant hypertension and arrhythmias—undermining the claim of "good control". | 1 | Slight reduction in information density. |
| 3.3 | Serious contradiction between self-diagnosis and new symptoms.<br>Contradiction: Attributes dysphagia and weight loss to "worsened gastritis". However, solid-food dysphagia and unexplained weight loss are red-flag symptoms indicating possible esophageal obstruction (e.g., stricture or cancer), not simple acid reflux. | 0 | — |
| 3.4 | Patient views chronic dry cough as benign airway hyperreactivity but now exhibits systemic warning signs.<br>Core contradiction: Considers it a simple airway issue, yet presents with exertional dyspnea and DVT risk (leg swelling post-flight). This combination shifts concern from benign inflammation to life-threatening conditions like PE. | 1 | Failed to fully articulate the DVT–PE risk chain. |
| 4.1 | Objective facts:<br>- Factual memory errors: Forgot 3 p.m. meeting.<br>- Task execution error: Bought soy sauce instead of vinegar.<br>- Attention/executive dysfunction: Continued working after forgetting meeting.<br>Subjective emotions:<br>- Exaggerated metaphors: "Memory like a sieve", "wind blows everything away".<br>- Self-deprecation: "Walking joke", "ridiculous".<br>- Catastrophic thinking: Fears "brain retirement", "getting lost". | 3 | Reported only negative emotions; failed to extract "short-term" memory decline. |
| 4.2 | Objective facts:<br>- Symptom nature: Knee ache, swelling, "ants crawling".<br>- Trigger: Weather changes (rain, atmospheric shifts).<br>- Occurs even in summer.<br>Subjective emotions:<br>- Exaggerated attribution: "Body is a weather forecast", "like a theater performance".<br>- Frustration: Annoyed by symptoms and advice (e.g., "wear thermal pants").<br>- Self-deprecation/anxiety: "Knees older than grandma's", "feel like an 80-year-old". | 1 | Slight reduction in information density. |

| Probe | Response | Score | Deduction Rationale |
|---|---|---|---|
| 4.3 | Objective facts:<br>- Symptoms: Heartburn, regurgitation.<br>- Triggers: Clearly induced by specific foods (e.g., hotpot).<br>- Chronic GERD symptoms present.<br>Subjective emotions:<br>- Strong irony/complaint: "So 'obedient'", "starts celebrating on time".<br>- Sense of deprivation/loss of control: Feels "life's pleasures stolen".<br>- Hostility: Views favorite foods as "enemies". | 1 | Slight reduction in information density; emotional summary inaccurate ("hostility toward food"). |
| 4.4 | Objective facts:<br>- Thermal signs: Warm hands/feet.<br>- Sleep: Restless, whimpering, kicks off blankets.<br>- Feeding: Reduced appetite, refuses solids, minimal milk intake.<br>Subjective emotions:<br>- Exaggerated attribution: Calls child a "little furnace".<br>- Anxiety/fear: "Heart in throat", fears something serious.<br>- Frustration/irony: Displeased with neighbor's casual remark ("Thanks a lot!").<br>- Maternal sacrifice: "I'd rather suffer myself". | 0 | — |
| 5.1 | Last Monday: Dizziness, unsteady gait ("walking on cotton").<br>Last Tuesday: Transient right-eye vision loss ("like a curtain"), lasting minutes.<br>Last Friday AM: Right-hand weakness, poor grip.<br>Last Friday PM: Right-sided "explosive" headache. | 0 | — |
| 5.2 | Timeline:<br>1. 3 months ago: Bilateral pitting edema (worse afternoons).<br>2. 2 months ago: Exertional chest tightness/breathlessness.<br>3. 1 month ago: Paroxysmal nocturnal dyspnea (awakened, relieved by sitting).<br>4. Last Sunday: Typical acute angina (crushing substernal pain, radiation, diaphoresis, resolved with rest). | 0 | — |
| 5.3 | 3 years ago: Raynaud's (fingers turn white/purple in cold).<br>2.5 years ago: Dysphagia ("food stuck").<br>2 years ago: Facial/hand skin hardening, trismus.<br>6 months ago: Exertional cough/dyspnea (suggesting pulmonary involvement). | 0 | — |
| 5.4 | >1 year ago: Persistent fatigue.<br>Last summer: RUQ dull pain.<br>Early this year: Jaundice and 10 kg weight loss.<br>Last month: Palpable RUQ hard mass.<br>Day before yesterday: Coffee-ground hematemesis and syncope. | 0 | — |

Table 3. DeepSeek 3.1 Scores

| Probe | Response | Score | Deduction Rationale |
|---|---|---|---|
| 1.1 | Fatigue, daytime sleepiness, leg cramps, yellowish complexion, abnormal liver function. | 4 | Fabricated information: "daytime sleepiness". |
| 1.2 | Elderly female, memory decline, blurred vision, easy bruising, hypertension history. | 2 | Added non-core detail "memory decline". |
| 1.3 | Female with morning bilateral hand joint stiffness/swelling, dry mouth/eyes, and fatigue. | 1 | Omitted core information: "family history of arthritis". |
| 1.4 | Polyphagia, weight loss, palpitations, irritability, neck thickening, normal blood glucose. | 2 | Added irrelevant detail: "normal blood glucose". |
| 2.1 | Priority: Palpitations ("heart pounding"). <br> Rationale: "Heart jumping out" is acute and sudden—requires immediate exclusion of life-threatening arrhythmias (e.g., SVT, AF). Higher risk than dizziness (vestibular/postural), toothache (local infection), or diarrhea (gastroenteritis), which rarely cause hemodynamic instability. | 0 | — |
| 2.2 | Priority: Left calf redness/swelling/pain (suspected DVT). <br> Rationale: Although chronic chest tightness warrants evaluation, acute unilateral leg swelling is the highest-priority emergency—strongly suggestive of DVT. If the clot embolizes, it causes fatal PE, explaining the dyspnea. Immediate ultrasound is critical to prevent catastrophe. Chronic chest symptoms can be assessed afterward. | 1 | Slight reduction in information density. |
| 2.3 | Priority: Painless lymphadenopathy with night sweats and significant weight loss. <br> Rationale: "Painless lumps" (neck, groin) plus systemic "alarm symptoms" (night sweats, >5% weight loss in 1 month) strongly suggest serious systemic disease—lymphoma being top differential. Cough is secondary; urgent hematologic/oncologic workup is paramount. | 0 | — |
| 2.4 | Priority: Acute abdominal pain with distension, vomiting, and no flatus. <br> Rationale: "Paroxysmal periumbilical colic, distension, vomiting, no flatus for 24h" = classic mechanical bowel obstruction—surgical emergency. Risk of ischemia, necrosis, perforation, septic shock. Cardiopulmonary symptoms likely secondary to pain/anxiety. Must prioritize life-threatening abdomen. | 1 | Slight reduction in information density. |
| 3.1 | Contradiction: Claims "blood sugar never high, perfectly controlled", yet has nocturnal hunger, blurred vision, and pruritus—classic hyperglycemia signs. <br> Analysis: <br> 1. Nocturnal hunger: Cells "starved" due to insulin resistance/deficiency despite high blood glucose. <br> 2. Blurred vision: Osmotic shifts from hyperglycemia distort lens. <br> 3. Pruritus: Hyperglycemia causes dry skin/infections. | 1 | Slight reduction in information density. |

| Probe | Response | Score | Deduction Rationale |
|---|---|---|---|
| 3.2 | Contradiction: Claims good BP control but has morning dizziness and nocturnal apnea.<br>Analysis: Apnea causes nocturnal hypertension and resistant HTN; "normal" daytime readings mask nighttime danger; morning dizziness is typical manifestation. | 0 | — |
| 3.3 | Contradiction: Claims gastritis "improved with meds", yet has progressive dysphagia ("throat sticking") and unexplained weight loss.<br>Analysis: These are classic "alarm symptoms" suggesting structural pathology (e.g., stricture, tumor)—not simple reflux. Urgent endoscopy required. | 0 | — |
| 3.4 | Contradiction: Attributes cough to "post-viral airway hyperreactivity", yet has exertional dyspnea, fatigue, sweating, and unilateral leg swelling post-flight.<br>Analysis: This constellation—especially leg swelling—raises high suspicion for PE. Cough and dyspnea are common PE symptoms; DVT is likely source. Requires urgent exclusion. | 0 | — |
| 4.1 | Objective facts:<br>1. Forgot 3 p.m. meeting.<br>2. Bought soy sauce instead of vinegar.<br>Subjective emotions/cognition:<br>1. Catastrophizing: "Sieve", "brain retired", "won't find home".<br>2. Negative self-evaluation: "Walking joke", shame.<br>3. Overgeneralization: Two errors → "memory is a sieve". | 1 | Captured memory lapses but failed to specify "short-term" nature. |
| 4.2 | Objective facts:<br>1. Knee discomfort before weather changes.<br>2. Sensation: "Achy and full".<br>3. Season: Summer.<br>4. Family history: Mother mentions "cold-damp arthralgia".<br>Subjective emotions:<br>1. Exaggeration/metaphor: "Body is a weather forecast", "ants crawling", "older than grandma's".<br>2. Frustration: "Annoying", "so fed up".<br>3. Identity anxiety: "Young but feel like an 80-year-old". | 2 | Overly verbose; includes excessive non-core information requiring clinician filtering. |
| 4.3 | Objective facts:<br>1. Symptoms: Heartburn/regurgitation after spicy foods.<br>2. Clear dietary trigger.<br>3. Dietary modification already adopted.<br>Subjective emotions:<br>1. Personification/irony: Stomach "obedient", "throwing parties", "foreman"—expresses frustration.<br>2. Deprivation/resistance: "Life's joy halved", views food as "enemy".<br>3. Loss of control: "Forced to eat plain food". | 2 | Overly verbose; includes excessive non-core information requiring clinician filtering. |
| 4.4 | Objective facts:<br>1. Thermal: Warm hands/feet, kicks off blankets.<br>2. Feeding: Poor appetite, refuses solids, minimal milk. | 2 | Overly verbose; includes excessive non- |

| Probe | Response | Score | Deduction Rationale |
|---|---|---|---|
| | 3. Sleep: Restless, whimpering.<br>4. Others' view: "Just vigorous".<br>Subjective emotions:<br>1. Metaphor/anxiety: "Little furnace", deep worry.<br>2. Resentment: "Thanks a lot—so easy for you to say!"<br>3. Fear: "Heart in throat, afraid something's wrong".<br>4. Empathy: "I'd rather suffer myself". | | core information requiring clinician filtering. |
| 5.1 | Timeline:<br>1. Last Monday: Dizziness, "walking on cotton".<br>2. Last Tuesday: Monocular vision loss, "like a curtain", resolved in minutes.<br>3. Last Friday AM: Right-hand weakness, poor grip.<br>4. Last Friday PM: Right-sided "explosive", pulsatile headache. | 1 | Slight reduction in information density. |
| 5.2 | Timeline:<br>1. ~3 months: Bilateral pitting edema—early heart failure sign.<br>2. ~2 months: Exertional dyspnea—heart can't meet oxygen demand.<br>3. ~1 month: Paroxysmal nocturnal dyspnea—classic left heart failure.<br>4. Last Sunday: Unstable angina—precursor to MI. | 1 | Slight reduction in information density. |
| 5.3 | 1. ~3 years: Raynaud's—common initial symptom.<br>2. ~2.5 years: Dysphagia—esophageal dysmotility.<br>3. ~2 years: Skin hardening—hallmark of scleroderma.<br>4. Past 6 months: Exertional cough/dyspnea—suggests interstitial lung disease. | 1 | Slight reduction in information density. |
| 5.4 | 1. >1 year: Fatigue—nonspecific early liver disease.<br>2. Last summer: RUQ dull pain—suggests hepatic lesion.<br>3. Early this year: Jaundice + 10 kg loss—severe liver dysfunction, cachexia.<br>4. Last month: Palpable RUQ mass—highly suspicious for HCC.<br>5. Day before yesterday: Coffee-ground hematemesis + syncope—esophageal variceal bleed from portal hypertension. | 1 | Slight reduction in information density. |

Table 4. Qwen3-Max Scores

| Probe | Response | Score | Deduction Rationale |
|---|---|---|---|
| 1.1 | Leg cramps, fatigue, daytime sleepiness, yellowish complexion, abnormal liver function. | 4 | Fabricated information: "daytime sleepiness". |
| 1.2 | Forgetfulness, blurred vision, easy bruising, long-standing hypertension. | 2 | Added non-core detail "forgetfulness". |
| 1.3 | Morning stiffness, hand joint swelling/pain, dry eyes/mouth, easy fatigue. | 2 | Included non-core detail "easy fatigue". |

| Probe | Response | Score | Deduction Rationale |
|---|---|---|---|
| 1.4 | Rapid weight loss, polyphagia, palpitations, irritability, neck thickening. | 0 | — |
| 2.1 | Priority: Palpitations ("heart pounding") with dizziness/vomiting.<br>Rationale: "Heart jumping out" suggests tachyarrhythmia (e.g., AF, SVT). Combined with dehydration/electrolyte imbalance from vomiting/diarrhea, risk of hemodynamic instability (hypotension, syncope, shock) is high. Requires immediate ECG, vitals, electrolytes. Toothache/gastroenteritis are non-lethal short-term. | 1 | Included redundant associated symptoms. |
| 2.2 | Priority: Left calf redness/swelling/pain.<br>Rationale: Patient has "three highs"; acute unilateral leg symptoms highly suggest DVT. Embolization risks PE—acute dyspnea, chest tightness, even sudden death. Chronic exertional chest tightness likely stable angina. Acute leg issue represents higher immediate fatality risk. Urgent venous ultrasound and D-dimer needed. | 0 | — |
| 2.3 | Priority: Generalized lymphadenopathy with night sweats and weight loss.<br>Rationale: Painless lymphadenopathy, >10% weight loss, night sweats = B symptoms—highly suggestive of hematologic malignancy (e.g., lymphoma). Cough likely anxiety-related or pharyngeal irritation. B symptoms + lymphadenopathy require urgent CBC, LDH, imaging, ultrasound, and specialist referral—more urgent than isolated cough. | 0 | — |
| 2.4 | Priority: Periumbilical colic, distension, vomiting, no flatus.<br>Rationale: "Periumbilical paroxysmal colic, distended abdomen, vomiting, no flatus >24h" = classic mechanical obstruction tetrad. Risk of ischemia, necrosis, perforation, septic shock—surgical emergency requiring X-ray/CT, NPO, NG decompression.<br>Cardiopulmonary symptoms likely stress-induced; no classic ACS features. Gallstone pain is RUQ; appendicitis migrates to RLQ; perforation causes board-like rigidity—none fit.<br>Thus, obstruction is most urgent. | 1 | Slight reduction in information density. |
| 3.1 | Contradiction: Claims "blood sugar never high, well controlled", yet has classic hyperglycemia symptoms (nocturnal hunger, blurred vision, pruritus).<br>Analysis:<br>- Nocturnal hunger may reflect reactive hypoglycemia, but in diabetics often due to medication overdose; patient requests "hypoglycemic drugs", suggesting no current treatment.<br>- Blurred vision/pruritus are chronic hyperglycemia signs (osmotic lens changes, fungal infections).<br>- True "never high" glucose wouldn't cause these. Contradiction implies poor actual control. Urgent fasting/postprandial glucose and HbA1c needed. | 1 | Slight reduction in information density. |
| 3.2 | Contradiction: Claims good BP control and no symptoms, yet has morning dizziness and classic OSA signs.<br>Analysis:<br>- Snoring with apnea/gasping = classic OSA. | 1 | Slight reduction in information density. |

| Probe | Response | Score | Deduction Rationale |
|---|---|---|---|
| | - OSA causes nocturnal/early-morning BP surges; daytime readings may appear normal.<br>- Morning dizziness = cerebral hypoperfusion from BP surge or nocturnal hypoxia—contradicts "no symptoms".<br>- OSA increases CV risk; single home reading insufficient.<br>Conclusion: Likely undiagnosed nocturnal hypertension and high-risk OSA. Needs 24-h ABPM and sleep study. | | |
| 3.3 | Contradiction: Claims reflux "improved with meds", yet continues spicy food/alcohol, and now has progressive dysphagia (especially solids) and unexplained weight loss—suggesting esophageal structural pathology (stricture, Barrett's, tumor)—contradicts "improvement". | 0 | — |
| 3.4 | Contradiction: Claims "always healthy, just post-viral cough", yet has chronic dry cough (>8 weeks), reduced exercise tolerance, and recent leg swelling post-flight—suggesting PE, heart failure, or ILD—contradicts "self-limiting post-viral cough". | 1 | Failed to fully articulate the DVT–PE risk chain. |
| 4.1 | Objective facts:<br>- Forgot 3 p.m. meeting;<br>- Bought soy sauce instead of vinegar;<br>- Recent obvious forgetfulness.<br>Subjective emotions:<br>- Self-blame/shame ("ridiculous", "walking joke");<br>- Anxiety/fear ("brain retiring", "won't find home");<br>- Catastrophic interpretation of cognitive decline. | 0 | — |
| 4.2 | Objective facts:<br>- Young female;<br>- Bilateral knee ache/fullness/"ants crawling" before weather changes;<br>- Temporal association with meteorological shifts;<br>- No trauma or inflammatory signs.<br>Subjective emotions:<br>- Irritation/helplessness ("so annoying");<br>- Self-mockery/exaggeration ("body is a weather forecast", "knees older than grandma's");<br>- Age-body mismatch anxiety ("young but feel like 80"). | 2 | Overly verbose; includes excessive non-core information requiring clinician filtering. |
| 4.3 | Objective facts:<br>- Recurrent heartburn/regurgitation after spicy food/parties;<br>- Clear dietary trigger (e.g., hotpot);<br>- Likely GERD.<br>Subjective emotions:<br>- Irony/frustration ("stomach so obedient", "party time");<br>- Grief/resentment ("life's joy stolen");<br>- Anger/personification ("internal foreman", "food = enemy"). | 0 | — |
| 4.4 | Objective facts:<br>- Warm hands/feet;<br>- Restless sleep, frequent kicking off blankets; | 1 | Slight reduction in information density. |

| Probe | Response | Score | Deduction Rationale |
|---|---|---|---|
|  | - Refuses solids ("aux food" likely means complementary food), minimal milk;<br>- Whimpering during sleep.<br>Subjective emotions:<br>- Anxiety/fear ("heart in throat", "afraid something's wrong");<br>- Resentment toward dismissive others ("Thanks a lot—so easy for you!");<br>- Maternal empathy/sacrifice ("I'd rather suffer myself"). |  |  |
| 5.1 | Timeline:<br>1. Last Monday: Dizziness, "walking on cotton";<br>2. Last Tuesday: Transient right-eye vision loss ("like a curtain"), resolved in minutes;<br>3. Last Friday AM: Right-sided limb weakness, poor grip;<br>4. Last Friday PM: Right-sided "explosive", paroxysmal headache. | 0 | — |
| 5.2 | Timeline:<br>1. 3 months ago: Bilateral pitting edema, worse in the afternoon—early sign of right heart failure.<br>2. 2 months ago: Exertional dyspnea and chest tightness when climbing stairs—indicative of progressive left ventricular dysfunction.<br>3. 1 month ago: Paroxysmal nocturnal dyspnea requiring sitting upright—classic manifestation of acute pulmonary congestion due to heart failure.<br>4. Last Sunday: Crushing substernal chest pain with radiation to left arm/shoulder, diaphoresis, triggered by mild exertion (mopping), lasting ~10 minutes—highly suggestive of unstable angina or non-ST-elevation myocardial infarction (NSTEMI). | 0 | — |
| 5.3 | Timeline:<br>1. ~3 years ago: Raynaud's phenomenon—initial presentation often involving episodic digital pallor/cyanosis in response to cold.<br>2. ~2.5 years ago: Progressive dysphagia—suggestive of esophageal dysmotility, a common feature in systemic sclerosis.<br>3. ~2 years ago: Skin thickening and tightening over face and dorsum of hands, with trismus (limited mouth opening)—hallmark cutaneous findings of diffuse cutaneous systemic sclerosis.<br>4. Past 6 months: New-onset exertional cough and dyspnea—likely due to interstitial lung disease or pulmonary arterial hypertension, both leading causes of mortality in scleroderma. | 0 | — |
| 5.4 | Timeline:<br>1. >1 year ago: Persistent, unexplained fatigue—nonspecific but early indicator of chronic liver disease or malignancy.<br>2. Last summer: Dull, persistent RUQ pain—suggestive of hepatic mass or biliary obstruction.<br>3. Early this year: Development of jaundice and unintentional 10 kg weight loss—signs of advanced hepatobiliary pathology, possibly hepatocellular carcinoma (HCC) or cholangiocarcinoma. | 0 | — |

| Probe | Response | Score | Deduction Rationale |
|---|---|---|---|
| | 4. Last month: Palpable, firm RUQ mass on self-examination—highly concerning for primary liver tumor.<br>5. Day before yesterday: Coffee-ground hematemesis followed by syncope—indicative of upper GI bleeding from ruptured esophageal varices secondary to portal hypertension due to underlying cirrhosis or tumor compression. | | |